\documentclass[10pt,twocolumn,letterpaper]{article}
\usepackage{iccv}              

\usepackage{colortbl}
\usepackage{pifont}
\usepackage{tikz}

\usepackage[dvipsnames]{xcolor}
\usepackage{enumitem}

\newcommand\rone[1]{\textbf{\textcolor{cyan}{WzrL}}}
\newcommand\rtwo[1]{\textbf{\textcolor{orange}{NtpB}}}
\newcommand\rthree[1]{\textbf{\textcolor{blue}{sTYB}}}

\usepackage{microtype}
\usepackage{graphicx}
\usepackage{booktabs} 
\usepackage{graphicx}
\usepackage{booktabs}
\usepackage{multirow}
\usepackage{adjustbox}
\usepackage{caption}
\usepackage{subcaption}
\usepackage{amsmath}

\definecolor{iccvblue}{rgb}{0.21,0.49,0.74}
\usepackage[pagebackref,breaklinks,colorlinks,allcolors=iccvblue]{hyperref}
\usepackage{amsmath}
\usepackage{amssymb}
\usepackage{mathtools}
\usepackage{amsthm}
\usepackage[capitalize]{cleveref} 

\theoremstyle{plain}

\theoremstyle{definition}

\theoremstyle{remark}

\usepackage[textsize=tiny]{todonotes}

\def\paperID{3047} 
\def\confName{ICCV}
\def\confYear{2025}

\title{Latent Diffusion Models with Masked AutoEncoders}

\author{%
  \makebox[\linewidth][c]{%
    Junho~Lee\textsuperscript{1}\footnotemark[1]\quad
    Jeongwoo~Shin\textsuperscript{1}\footnotemark[1]\quad
    Hyungwook~Choi\textsuperscript{1}\quad
    Joonseok~Lee\textsuperscript{1}\footnotemark[2]%
  }\\[0.5em]
  \textsuperscript{1}Seoul National University, Seoul, Korea\\
  {\tt\small \{joon2003,swswss,chooi221,joonseok\}@snu.ac.kr}
  \vspace{-0.5em}
}

\begin{document}
\renewcommand\thefootnote{\fnsymbol{footnote}}
\twocolumn[{
    \maketitle
    \begin{center}
        \includegraphics[width=\textwidth]{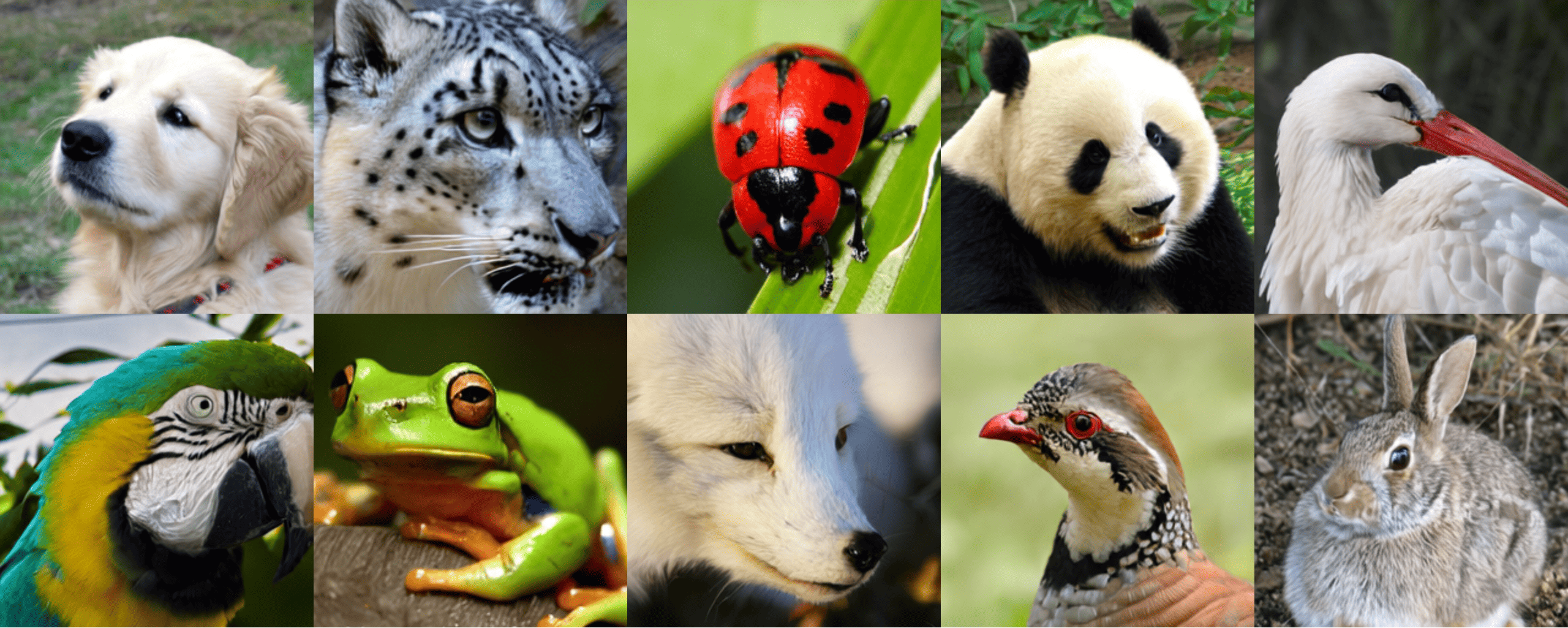}
        \captionof{figure}{
        \textbf{Class-conditional generation samples of our proposed LDMAE.} Trained on ImageNet-1K, resolution of $256 \times 256$.}
        \vspace{1.0em}
        \label{fig:dm_teaser}
    \end{center}
}]
\footnotetext[1]{Equal contribution.}
\footnotetext[2]{Corresponding author.}
\renewcommand\thefootnote{\arabic{footnote}}

\begin{abstract}
\label{sec:abstract}
In spite of the remarkable potential of Latent Diffusion Models (LDMs) in image generation, the desired properties and optimal design of the autoencoders have been underexplored.
In this work, we analyze the role of autoencoders in LDMs and identify three key properties: latent smoothness, perceptual compression quality, and reconstruction quality.
We demonstrate that existing autoencoders fail to simultaneously satisfy all three properties, and propose Variational Masked AutoEncoders (VMAEs), taking advantage of the hierarchical features maintained by Masked AutoEncoders.
We integrate VMAEs into the LDM framework, introducing Latent Diffusion Models with Masked AutoEncoders (LDMAEs).
Through comprehensive experiments, we demonstrate significantly enhanced image generation quality and computational efficiency.
Our code is available at https://github.com/isno0907/ldmae.
\end{abstract}


\section{Introduction}
\label{sec:intro}
Diffusion models (DMs)~\cite{ho2020ddpm, song2020score, karras2022edm, dhariwal2021diffusion, nichol2021improved, peebles2023dit} have recently set new standards in image generation, but their iterative denoising process incurs high computational cost. While deterministic sampling~\cite{song2020ddim, lu2022dpm} reduces the number of sampling steps, the core inefficiency of operating directly in pixel space remains.
To address these challenges, Latent Diffusion Models (LDMs)\cite{rombach2022ldm} shift the denoising process from pixels to a compressed space, focusing on semantic representations with lower computational cost. For this, Variational Autoencoders (VAEs)\cite{kingma2013vae} are adopted to enforce probabilistic alignment with a prior, enabling smooth interpolation, generative sampling, and robust generalization.


However, this latent space alignment loss
often negatively impacts the reconstruction performance, leading to a trade-off between latent space alignment and reconstruction quality.
To mitigate this, LDMs, such as Stable Diffusion~\cite{rombach2022ldm,esser2024sd3},
incorporate adversarial and perceptual losses to further improve the VAEs, which is commonly referred to as the StableDiffusion-VAEs (SD-VAEs).
Subsequent studies further improved it by enhancing reconstruction quality via increased latent dimensionality~\cite{esser2021vqgan, flux2023, dai2023emu, vavae} or by improving efficiency with more lightweight architectures~\cite{sadat2024litevae}.

\begin{figure}[t]
    \vspace{-0.3cm}
    \centering
    \includegraphics[width=0.75\linewidth]{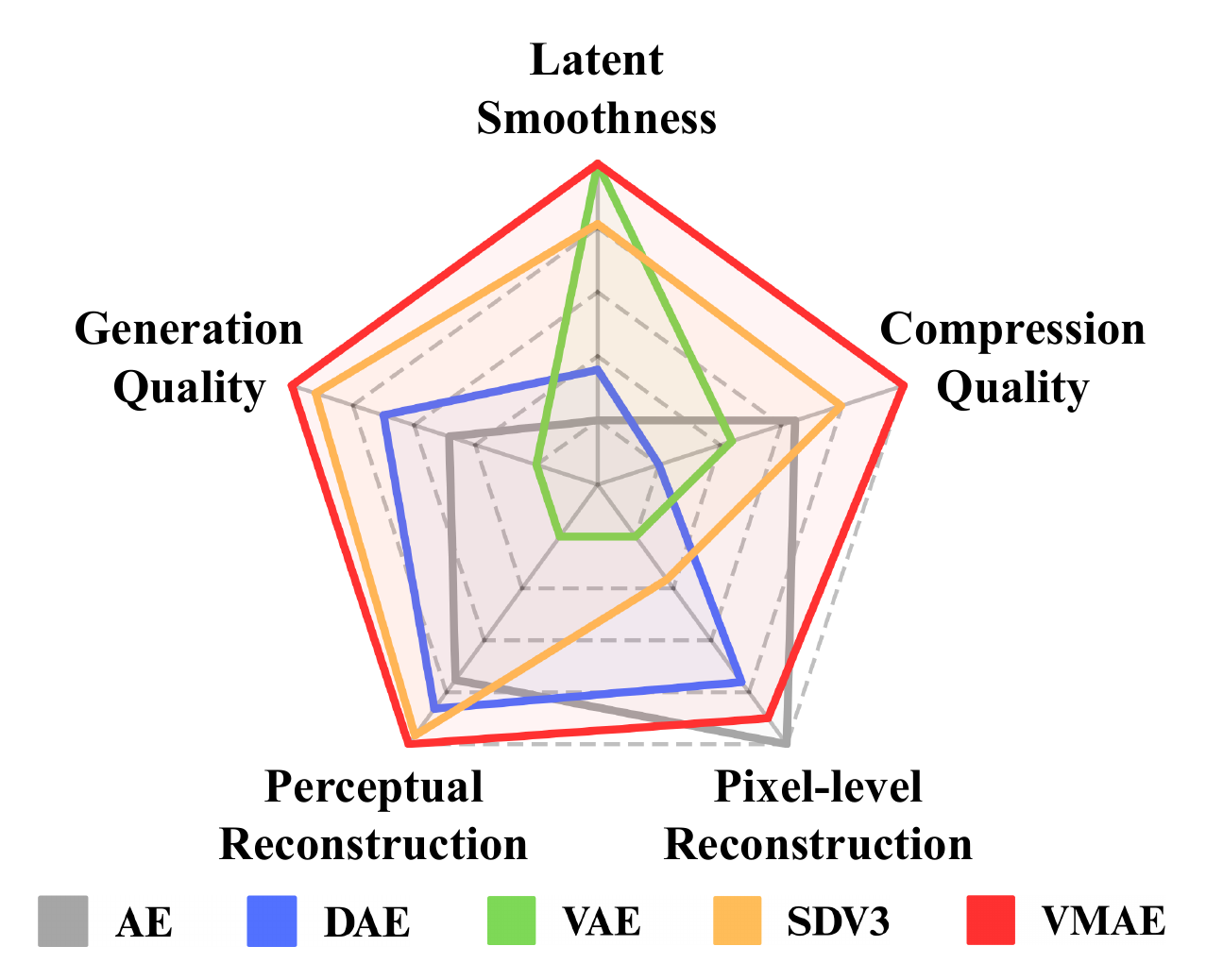}
    \caption{Comparison of autoencoding methods 
    across key metrics: latent smoothness, perceptual compression quality, reconstruction quality (at pixel and perceptual levels),
    and overall generation quality. Each score is linearly rescaled so that the outermost and innermost grids indicate the highest and lowest scores, respectively.}
    \label{fig:intro:ae_comparison}
    \vspace{-0.5cm}
\end{figure}

Despite the notable success of the additional loss terms in autoencoders, their specific roles in shaping the characteristics of autoencoders, as well as their impact on the entire LDM framework, remain underexplored.
A fundamental question arises: what properties should an autoencoder possess to effectively support LDMs?
Since the choice of autoencoder directly influences the quality and efficiency of LDMs, this is a crucial question for designing an autoencoder. To address this question, we propose three key properties that an autoencoder should satisfy to effectively compress data, preserve essential semantics, and eventually facilitate high-quality image generation in the LDM framework: 1) latent smoothness, 2) perceptual compression quality, and 3) reconstruction quality.

First, \emph{latent smoothness} ensures that minor perturbations in the latent representation do not drastically change the generated output.
A sufficiently smooth latent space allows the model to generate high-quality images robustly, even when the diffusion model predicts latent codes with some error.
We reveal through an empirical analysis on the stability of autoencoder embeddings under small perturbations (\cref{sec:exp:ae_analysis}) that
autoencoders with deterministic encoding, such as vanilla AutoEncoders (AEs), lack this property, whereas probabilistic encoding, as used in VAEs, much better satisfies this requirement.

Second, \emph{perceptual compression quality} measures how effectively an autoencoder encodes the images into a simplified latent space, maximizing the reduction of perceptual details while preserving semantics as much as possible. However, the boundary between `semantics' and `perceptual details' is inherently ambiguous, as they exist on a continuous spectrum. This spectrum ranges from pixel-level (minimal compression) to object-level (maximum compression).
In \cref{sec:exp:ae_analysis}, we observe that SD-VAEs aggressively compress input images, resulting in compression close to the object-level. This simplifies the latent space, making it easier for diffusion models to predict latent codes, but sacrifices fine-grained details. To mitigate this, we propose \emph{hierarchical compression}, which organizes the latent space into a continuous hierarchy: first grouping images into object-level clusters, then subdividing each object into part-level clusters (e.g., facial features), and finally refining them into fine-detail clusters. This structure preserves global simplicity while retaining local details, effectively balancing compression and reconstruction quality.

Lastly, \emph{reconstruction quality} ensures the decoder to accurately reconstruct the original image from its latent representation.
We claim that this property needs to be assessed both at the perceptual level (\emph{e.g.}, measured by rFID or LPIPS) and at the pixel level (\emph{e.g.}, evaluated by PSNR and SSIM).
In literature, perceptual metrics such as FID are widely adopted, while pixel-level reconstruction quality has been overlooked, potentially due to the lack of reference images.
Nevertheless, ensuring high pixel-level reconstruction quality is crucial, particularly for tasks that require precise spatial or numerical consistency, such as image editing, inpainting, or super-resolution.
Our empirical analysis reveals that SD-VAEs achieve high reconstruction quality at perceptual level, while their pixel-level reconstruction quality is notably weaker.

From systematical assessments of these properties (see \cref{sec:exp:ae_analysis} for details), we provide a concise summary of various autoencoders in \cref{fig:intro:ae_comparison}.
We clearly observe that no existing autoencoders satisfy all of these desired properties, lacking at least one of them.
Inspired by the recent findings of masked autoencoding~\cite{shin2024selfmae} that it constructs a well-aligned hierarchical feature space, we propose a novel transformer-based \emph{Variational Masked AutoEncoders (VMAEs)}.
VMAE achieves both latent smoothness and high reconstruction quality through probabilistic alignment, tailored loss design, and structural enhancements, while retaining the hierarchical compression benefits of MAEs.
Additionally, VMAE significantly improves computational efficiency, using only 13.4\% of the parameters and 4.1\% of the GFLOPs of SD-VAE. It also converges 2.7 times faster with greater training stability.

Furthermore, we incorporate our VMAEs into the LDMs framework, referred to as Latent Diffusion Models with Masked AutoEncoders (LDMAEs).
Through a series of experiments, we verify that LDMAEs improve generative performance across multiple datasets and settings.
These findings support our hypothesis that the three key properties contribute to effective latent diffusion.

We summarize our contributions as follows:
\begin{itemize}[leftmargin=5mm]
    \setlength{\itemsep}{0pt}
    \setlength{\parskip}{0pt}
    \item We identify smooth latent space, effective perceptual compression, and high reconstruction quality as desired properties of autoencoders for LDMs with empirical evidence from thoroughly designed experiments.
    \item Based on our analysis, we introduce VMAEs, a novel autoencoder designed to improve all three key factors with significant model efficiency.
    \item Through extensive experiments, we demonstrate superior generation performance of our LDMAE, equipped with our VMAE as its autoencoder.
\end{itemize}

\section{Related Work}
\label{sec:related}


\textbf{Diffusion Models.}
Diffusion models generate images by reversing a forward noise process. DDPM~\cite{ho2020ddpm} introduces a Markovian framework where Gaussian noise is incrementally added, and a neural network learns to denoise by optimizing Evidence Lower Bound, but requires many sampling steps. 
Score-based models~\cite{song2020score} generalize diffusion to continuous time, enabling efficient solvers via reverse-time stochastic differential equations. ADM~\cite{dhariwal2021diffusion} improves scalability with architectural refinements and classifier guidance.

These standard diffusion models~\cite{ho2020ddpm, dhariwal2021diffusion, song2020ddim} operate in pixel space, which presents significant computational challenges, especially for high-resolution sampling. To address this, Latent Diffusion Models (LDMs)~\cite{rombach2022ldm, peebles2023dit} shift the diffusion process to a learned latent space, using a pre-trained autoencoder to represent data in a perceptually compressed form, removing redundant pixel-level details.

\noindent
\textbf{Autoencoders for Image Generation.}
Autoencoders~\cite{10.5555/104279.104293, vincent2008dae, kingma2013vae, he2022mae} encode input data into a compact latent space that preserves structure, reducing computational overhead and enabling tasks like interpolation and reconstruction.

Recent studies have explored integrating autoencoder architectures into image generation. Two main approaches are based on on MAEs~\cite{he2022mae} and VAEs~\cite{kingma2013vae}. MAEs-based methods, like MaskGIT~\cite{chang2022maskgit} and MAGE~\cite{li2023mage}, adapt masked encoding to discrete token-based space, while MAR~\cite{li2024autoregressive} utilizes diffusion to generate continuous tokens. VAEs-based methods sample images from a Gaussian distribution.  VQVAE~\cite{van2017neural} and VQGAN~\cite{esser2021vqgan} further employ discrete latent spaces with a codebook for improved generation.

LDMs~\cite{rombach2022ldm}, however, directly generate compressed latents from autoencoders, which can be decoded back into images, accelerating generation but often degrading reconstruction quality. To address this, StableDiffusion3~\cite{esser2024sd3}, Emu~\cite{dai2023emu}, and FLUX~\cite{flux2023} introduce larger latent dimensionality to capture finer details. VA-VAE~\cite{vavae} adds alignment losses with vision foundation models, and DC-AE~\cite{chen2024deep} aims for high spatial compression without sacrificing fidelity.

\section{Preliminary}
\label{sec:prelim}

\subsection{Autoencoders}
\label{sec:prelim:autoencoder}

We compare four classical autoencoder variants within LDMs~\cite{rombach2022ldm} and include MAEs~\cite{he2022mae}, the foundation of our model. See \cref{apdx:prelim:autoencoders} for details.

The vanilla \textbf{AutoEncoders (AEs)}~\cite{10.5555/104279.104293} are optimized solely by the reconstruction loss, compressing the input into a more compact latent space and reconstructing it back. 
\textbf{Denoising AutoEncoders (DAEs)}~\cite{vincent2008dae} extend AEs by introducing noise to the input data and training the model to recover the original noise-free input for robust latent features with improved generalization.
\textbf{Variational AutoEncoders (VAEs)}~\cite{kingma2013vae} adopt a probabilistic framework,
encoding input data into a latent distribution instead of a fixed vector, with KL-divergence regularizing it towards a Gaussian distribution to enable sampling.
\textbf{StableDiffusion VAEs (SD-VAEs)}~\cite{rombach2022ldm,esser2024sd3} 
are built upon VQGAN~\cite{esser2021vqgan}, the autoencoder adopted in traditional LDMs~\cite{rombach2022ldm}.
Following the VQGAN, SD-VAEs integrate an additional adversarial network and train with perceptual loss for an improved perceptual quality. Unlike VQGAN, however, SD-VAEs omit the quantization layer entirely and instead use continuous features.
In this paper, we follow the SD-VAEs settings from StableDiffusion3~\cite{esser2024sd3}, the state-of-the-art LDM, unless noted otherwise.

\textbf{Masked AutoEncoders (MAEs)}~\cite{he2022mae} were introduced as a self-supervised representation learning method using Vision Transformers (ViTs)~\cite{dosovitskiy2020vit}. The encoder maps a randomly masked-out image to fixed vectors, while the decoder predicts the masked regions using these latent vectors along with learnable mask tokens.


\subsection{Latent Diffusion Models}
\label{sec:prelim:ldm}

Standard DMs~\cite{ho2020ddpm,dhariwal2021diffusion,song2020ddim} operate in pixel space $\boldsymbol{x} \sim p_{\text{data}}$ and optimize the following objective:
\begin{equation}
\mathcal{L}_{\text{DM}} = \mathbb{E}_{\boldsymbol{x}, \epsilon \sim \mathcal{N}(0,1), t} \left[ \left\| \epsilon - \epsilon_\theta \big( \boldsymbol{x}_t, t \big) \right\|_2^2 \right],\nonumber
\end{equation}
where $\epsilon_\theta$ is the model that predicts the original noise $\epsilon$ given the corrupted input \( \boldsymbol{x}_t \) at timestep $t$.



LDMs~\cite{rombach2022ldm} extend this framework to the latent space $\boldsymbol{z} \sim \mathcal{E}(\boldsymbol{x})$ encoded by an encoder $\mathcal{E}$, instead of pixel space:
\begin{align} 
    \! \mathcal{L}_{\text{LDM}} &= \mathbb{E}_{\mathcal{E}(\boldsymbol{x}), y, \epsilon \sim \mathcal{N}(0,1), t} \left[ \left\| \epsilon - \epsilon_\theta \big( \boldsymbol{z}_t, t, \tau_\theta(y) \big) \right\|_2^2 \right],\nonumber
\end{align}
where $\tau_\theta(\boldsymbol{y})$ projects the conditioning inputs $\boldsymbol{y}$ into the latent space, enabling multi-modal generative tasks.


\section{Variational Masked AutoEncoders}
\label{sec:method}

We introduce our novel autoencoder, called \emph{Variational Masked AutoEncoders (VMAEs)}, which embody the three desirable properties for LDM framework, \emph{i.e.}, smooth latent space (\cref{method:smoothness}), perceptual compression (\cref{method:compression}), and high reconstruction quality (\cref{method:reconstruction}).

\begin{figure}[t]
    \centering
    \includegraphics[width=\linewidth]{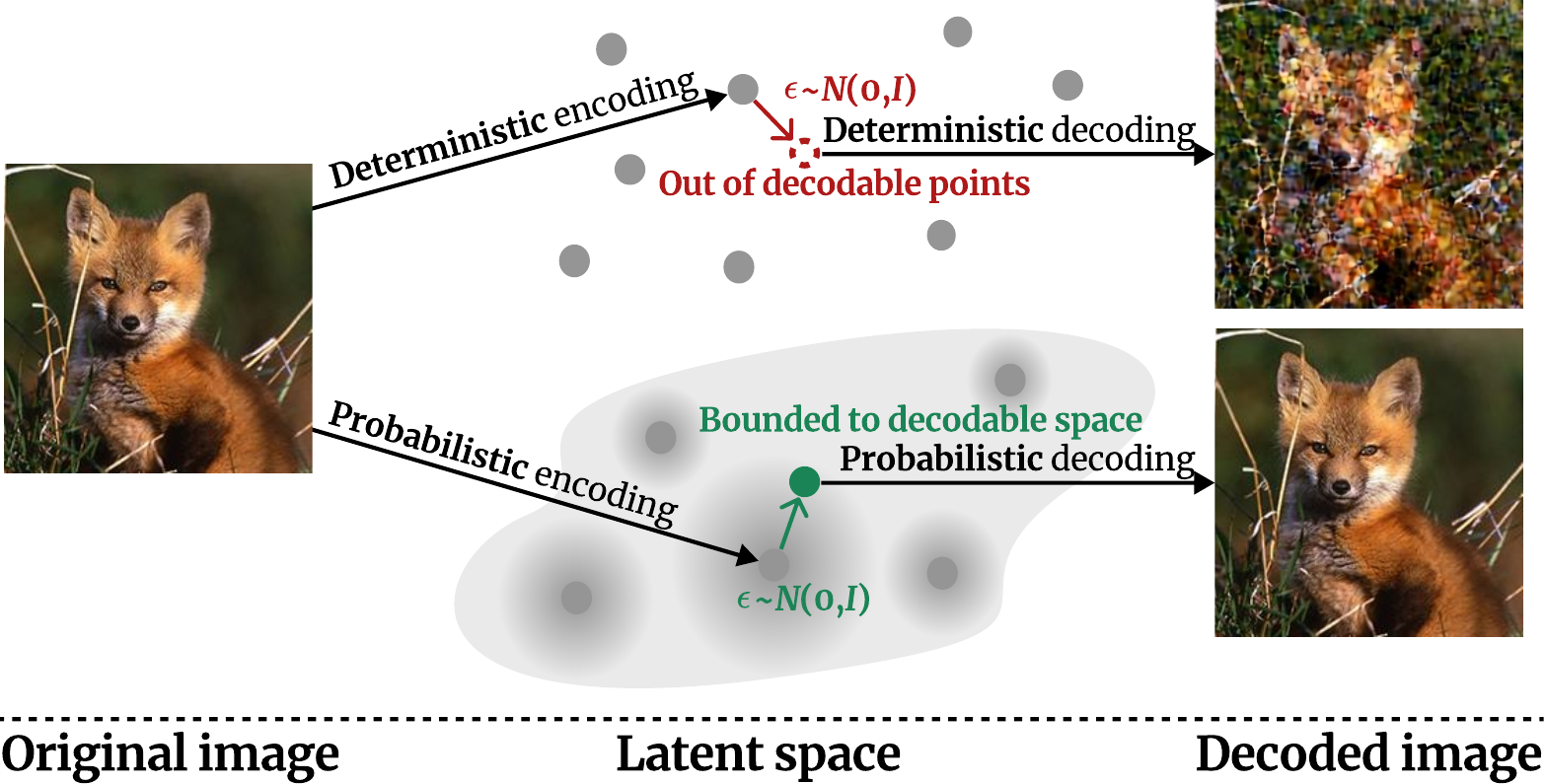}
    \caption{\textbf{Sparse latent points \emph{vs.} smooth latent space.} 
    Deterministic decoders allow only the exact points to be \emph{decodable} due to its \emph{sparse latent points}. In contrast, probabilistic decoders, with a \emph{smooth latent space}, ensure that all latent points in the vicinity of an encoded point remain within the \emph{decodable} space.}
    \vspace{-.4cm}
    \label{fig:support}
\end{figure}

\subsection{Support of Latent Space}
\label{method:smoothness}
We emphasize that \emph{probabilistic latent space} is crucial for training LDMs, as it ensures a \emph{smooth} latent space—a fundamental requirement of the diffusion process.
An overly sparse or low-dimensional latent manifold can hinder diffusion modeling by causing inconsistent score matching~\cite{choi2021not, song2019generative}.

Intuitively, when the target space is sparse, the diffusion model needs to learn \emph{exact} point mappings, making training highly challenging.
In LDMs, latent sparsity is closely tied to decoder capability—if the decoder can reconstruct only from a limited set of discrete points, the diffusion model is constrained to generating those exact points for successful reconstruction.
In contrast, in a \textit{fully-supported} latent space—\emph{i.e.}, where the decoder reliably reconstructs from any point—the diffusion model is allowed to flexibly generate latents within a vicinity of the target point, enabling a more robust generation process.


This phenomenon is illustrated in \cref{fig:support}, comparing deterministic autoencoders (AEs, DAEs) with probabilistic ones (VAEs, SD-VAEs, VMAEs).
Deterministic models encode inputs into fixed vectors, creating a sparse latent space where only exact encodings are \emph{decodable}, while most other points remain inaccessible. Even slight perturbations by Gaussian noise make decoding difficult. In contrast, probabilistic ones establish a continuous latent space, ensuring that points within a noise-induced vicinity remain \emph{decodable}, enabling stable reconstruction. See \cref{sec:exp:ae_analysis} for a detailed analysis.

Reflecting this observation, we design our VMAEs to encode input data $\boldsymbol{x} \sim p_{\text{data}}(x)$ as probabilistic distributions $q_\phi(\boldsymbol{z}|\boldsymbol{x})$ rather than fixed points, promoting smoother latent space.
Additionally, we regularize the latent distribution toward a Gaussian prior $p(\boldsymbol{z})$ via KL divergence, ensuring strict adherence to the Variance Preserving (VP) probability path condition~\cite{song2020score}, \emph{i.e.}, unit variance, by minimizing 
\begin{align}
  \mathcal{L}_{\text{reg}} = \mathbb{E}_{p_{\text{data}}(\boldsymbol{x})} \Big[{\text{D}_\text{KL}(q_\phi(\boldsymbol{z}|\boldsymbol{x}) | p(\boldsymbol{z}))} \Big]. 
  \label{eq:loss_kl}
\end{align}

\subsection{Perceptual Compression}
\label{method:compression}

Following the original LDMs~\cite{rombach2022ldm},
\emph{perceptual compression} refers to encoding raw data into a compact latent space where fine-grained perceptual details are discarded, while preserving macroscopic semantics. 
Then, the decoder is responsible for recovering these details; that is,
an ideal autoencoder encodes semantically similar features identically in the latent space, and the decoder restores the distinctiveness
by adding specific perceptual details back.

However, it is inherently impossible to precisely define the boundaries between `semantics' (to be preserved) and `details' (to be compressed).
Rather, these aspects lie on a continuous spectrum, ranging from the lowest pixel-level (no compression) to the most abstract object-level (most compressed).
Intermediate levels can include parts (e.g., eyes or nose) or patterns (e.g., texture or color). Given this, we claim that it is crucial to design an autoencoder with a compression level optimized for LDM framework.



\begin{figure}[t]
    \centering
    \begin{minipage}[c]{0.2\linewidth}  
        \centering
        \raisebox{0.0\height}{%
            \includegraphics[width=\linewidth]{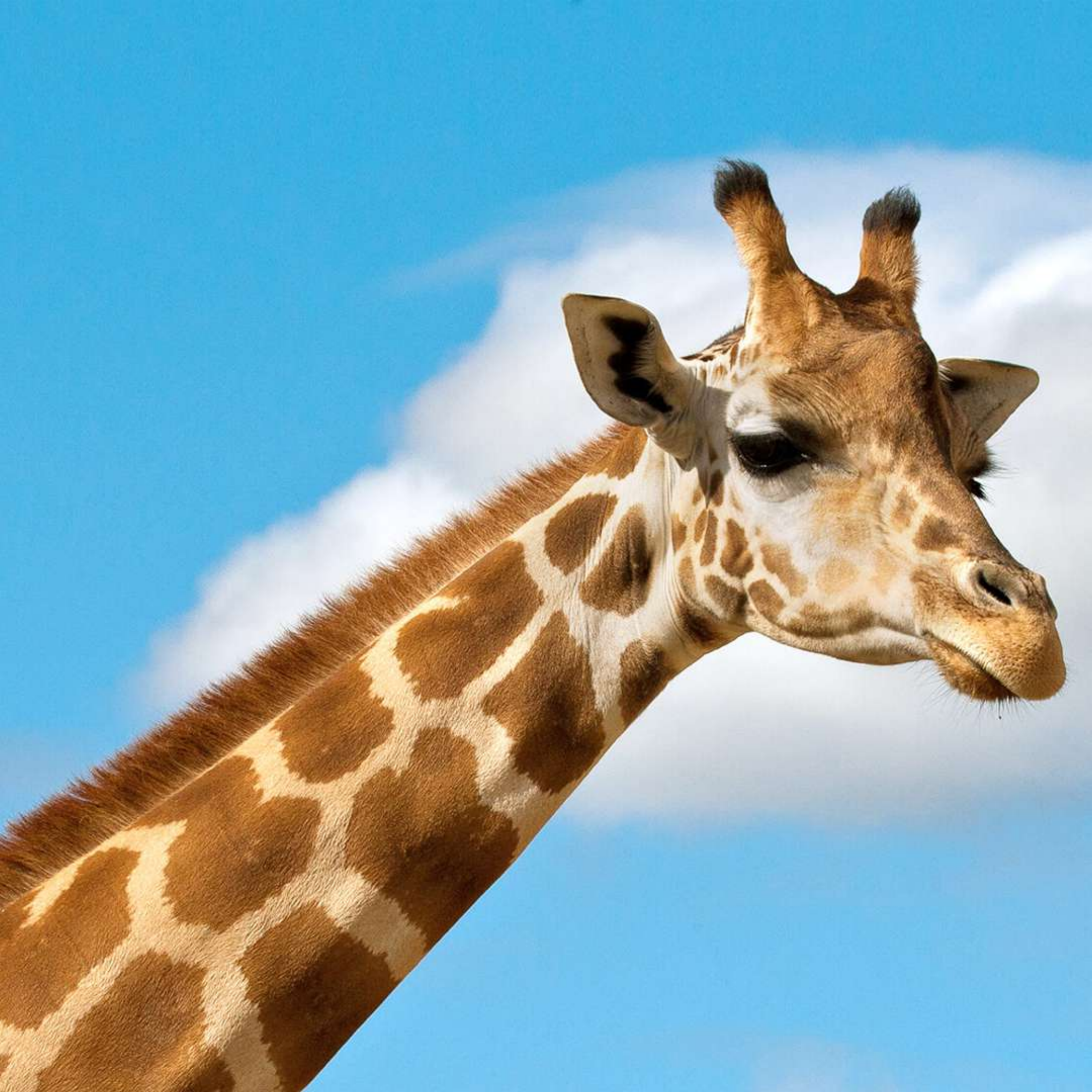}
        }
        \vspace{-1mm}  
        \centering
        {\small Input image}
    \end{minipage}
    \hfill
    \begin{minipage}[c]{0.75\linewidth} 
        \centering
        \includegraphics[width=\linewidth]{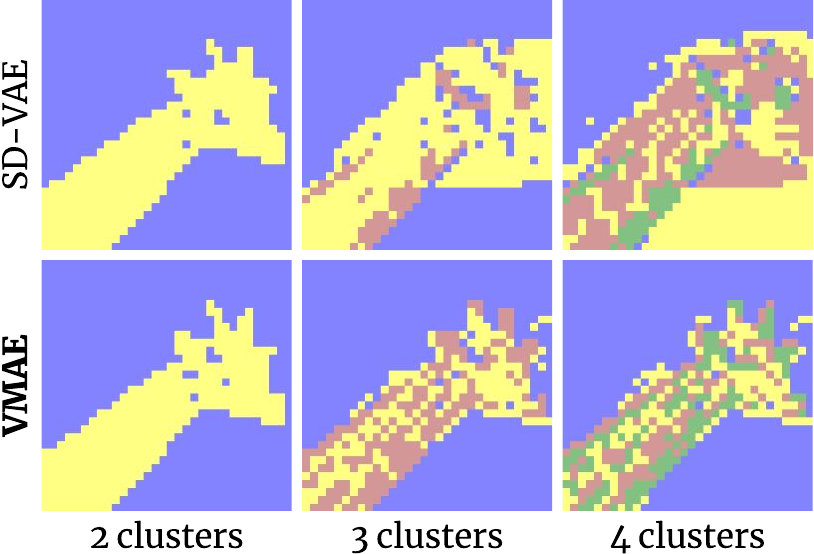}
    \end{minipage}
    \caption{\textbf{Separability of latent features.} Our VMAE achieves a more highly separable latent space while maintaining strong semantic grouping, indicating \emph{hierarchical} compression. In contrast, SD-VAE primarily exhibits semantic clustering without clearly differentiating features within each cluster.}
    \vspace{-.4cm}
    \label{fig:perceptual_compression}
\end{figure}

\cref{fig:perceptual_compression} (top) illustrates unsupervised clustering of the local features of SD-VAE for a giraffe image, with various number of clusters.
While SD-VAE clearly distinguishes the main object (giraffe) from the background (2 clusters), it struggles to capture finer details like fur patterns (4 clusters).
This suggests that its compression ability is limited to the highest object-level, renouncing to represent finer details in the latent space.
This simplified latent space can facilitate the learning process of the diffusion model~\cite{chen2025maetok}, but it may result in the loss of perceptual details at decoding, as most fine details are discarded during the encoding process.
Also, as the fine-grained features are entangled in the latent space, downstream tasks like image editing, where disentangled features are particularly critical~\cite{nam2024contrastive, hertz2023delta, shuai2024latent, hahm2024isometric} would not be performed well.
Based on these intuitions, we explore an alternative autoencoder with optimal perceptual compression.

Recent studies~\cite{park2023sslanalysis, shin2024selfmae} have demonstrated that the encoded features of MAEs are highly distinguishable based on the perceptual visual patterns such as texture and color,
learned through its \emph{masked-part prediction} objective.
Moreover, the latest analysis on MAEs~\cite{shin2024selfmae} further confirms that
this training strategy leads the encoded patch vectors to be \emph{hierarchically} clustered in the embedding space from the abstract objects to simpler visual patterns.

These findings pave the way to equip an autoencoder with \emph{continuously hierarchical compression} 
by leveraging MAEs. That is, latent features are first compressed at the object-level, then progressively diversified down to the pattern-level hierarchically, with the degree of clustering decreasing at each stage.
\cref{fig:perceptual_compression} (bottom) illustrates that our MAEs-based latent features are gradually distinguishable from high-level objects (2 clusters) to finer fur patterns (4 clusters).
showcasing its hierarchical compression.
This hierarchical compression meets the needs of both autoencoders and LDMs: 1) strongly clustered features at higher levels simplify diffusion model training, and 2) discriminative features across various levels create a highly disentangled, yet hierarchically structured 
latent space, enabling high reconstruction quality.

Considering these aspects, we incorporate the masked-part prediction objective for hierarchical perceptual compression.
Specifically, the probabilistic decoder $p_\theta$ is trained to predict the masked regions $\boldsymbol{x}_m$, given only the latent variables of the visible regions, $\boldsymbol{z} \sim q_{\phi}(\boldsymbol{z}|\boldsymbol{x}_v)$, by optimizing:
\begin{align}
  \mathcal{L}_{\text{M}} = \mathbb{E}_{p_{\text{data}}(\boldsymbol{x}), p(\boldsymbol{x}_v|\boldsymbol{x}), p(\boldsymbol{x}_m|\boldsymbol{x}),q_{\phi}(\boldsymbol{z}|\boldsymbol{x}_v)}[-\log p_\theta(\boldsymbol{x}_m|\boldsymbol{z})]. \nonumber
\end{align}
As $\boldsymbol{z} \sim q_{\phi}(\boldsymbol{z}|\boldsymbol{x}_v)$ in masked-part prediction, the regularization for the latent distribution in Eq.~\eqref{eq:loss_kl}  is modified to
\begin{align}
  \mathcal{L}_{\text{reg}} = \mathbb{E}_{p_{\text{data}}(\boldsymbol{x}), p(\boldsymbol{x}_v|\boldsymbol{x})} \Big[{\text{D}_\text{KL}(q_\phi(\boldsymbol{z}|\boldsymbol{x}_v) | p(\boldsymbol{z}))} \Big].\nonumber
  \label{eq:loss_kl2}
\end{align}



\subsection{Perceptual Reconstruction}
\label{method:reconstruction}

\begin{figure}[t]
    \centering
    \includegraphics[width=\linewidth]{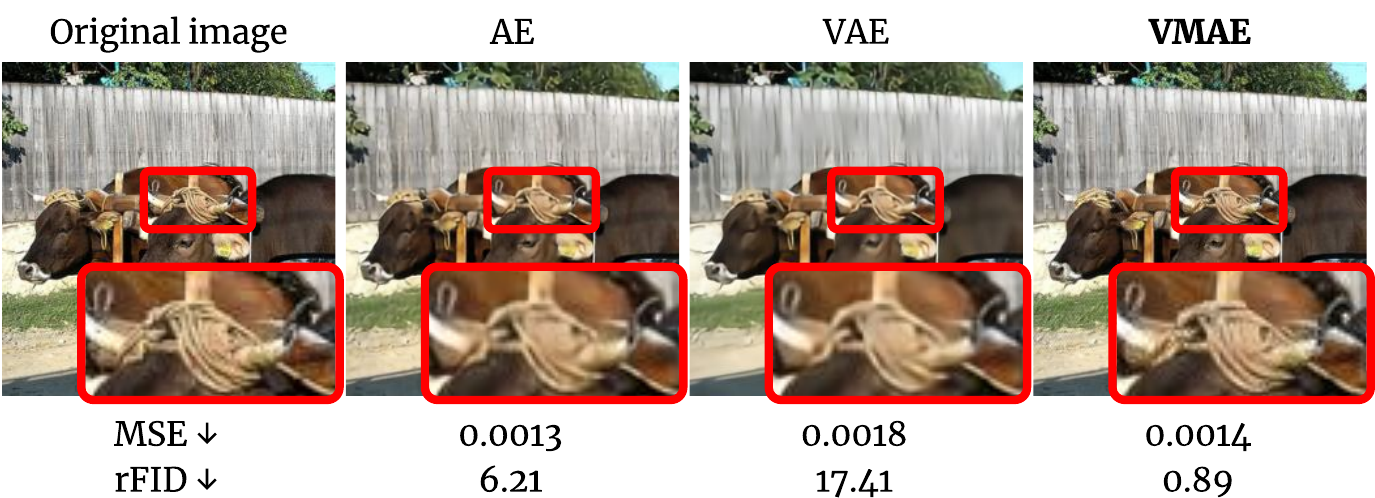}
    \vspace{-0.5cm}
    \caption{\textbf{Key factors for reconstruction quality.} Although all autoencoders achieve a comparable pixel-level reconstruction (in MSE), they differ in perceptual details especially within the boxed region (in rFID), suggesting that the autoencoders should be optimized for both pixel-level accuracy and perceptual restoration.}
    \vspace{-.4cm}
    \label{fig:reconstruction}
\end{figure}

Although autoencoders solely trained to minimize the pixel-wise distance (\emph{e.g.}, AEs) demonstrate strong reconstruction capability, they do not necessarily recover perceptual details in the pixel space.
As illustrated in \cref{fig:reconstruction}, AE achieves an optimal reconstruction with a very low MSE.
However, closer inspection of the boxed region reveals subtle differences, such as the lumped details of the ropes on the cow's head, which affect perceptual fidelity.
For VAE, neither pixel accuracy nor perceptual fidelity is achieved, resulting in blurry reconstruction, losing most perceptual details.
To address the loss of perceptual details in decoding, we train our VMAEs with both perceptual and reconstruction losses.

Since the reconstruction loss can be applied only to the visible regions $\boldsymbol{x}_v$ in the masked-part prediction scheme,
\begin{equation}
  \mathcal{L}_{\text{R}} = \mathbb{E}_{p_{\text{data}}(\boldsymbol{x}), p(\boldsymbol{x}_v|\boldsymbol{x}), q_{\phi}(\boldsymbol{z}|\boldsymbol{x}_v)}[-\log p_\theta(\boldsymbol{x}_v|\boldsymbol{z})].\nonumber
  \label{eq:loss_recon}
\end{equation}


For perceptual loss, we use LPIPS~\cite{zhang2018lpips} loss, implemented as a reconstruction loss at the feature level:
\begin{equation}
  \mathcal{L}_{\text{P}} = {\mathbb{E}_{p_{\text{data}}(\boldsymbol{x}), q_\phi(\boldsymbol{z}|\boldsymbol{x}_v), p_\theta(\hat{\boldsymbol{x}}|\boldsymbol{z})} \bigg[ \sum_{l} w_l \|\psi_l(\boldsymbol{x}) - \psi_l(\hat{\boldsymbol{x}})\|_2^2 \bigg]}, \nonumber
\end{equation}
where $\psi_l$ extracts features up to the $l$-th layer of a pre-trained network (\emph{e.g.}, VGG~\cite{simonyan2014vgg}), with $w_l$ as its layer weight.


\subsection{Model Architecture and Training Objective}

Putting all ideas together, we finally propose the model architecture and training objective of our proposed method, Variational Masked AutoEncoders (VMAEs) in detail.

\vspace{0.1cm} \noindent
\textbf{Model Architecture.}
We adopt a symmetric ViT-based encoder and decoder.
At encoding, an image is spatially
downsampled by patchification
and fed into Transformer layers.
The encoded features are then reduced to a pre-defined size.
With the symmetric structure, the decoding process mirrors the encoding process in reverse order.

\vspace{0.1cm} \noindent
\textbf{Training Objective.}
Based on our observations of latent space, we claim that the three aforementioned properties are essential for autoencoders to effectively support LDMs.
To satisfy these, we design our encoder and decoder as probabilistic functions to construct a smooth latent space, regularized with a KL divergence to satisfy the VP condition (\cref{method:smoothness}).
To preserve discriminative features essential for hierarchical compression~\cite{shin2024selfmae}, we allow the predicted mean to be learnable without enforcing it to zero, 
while still regularizing the variance to be unit.
More importantly, we incorporate the masked-part prediction loss to achieve hierarchical compression (\cref{method:compression}) and the perceptual loss for perceptually accurate reconstruction (\cref{method:reconstruction}).
Overall, the final training objective of our VMAE is

\begin{equation}
  \mathcal{L}_{\text{VMAE}} = \mathcal{L}_{\text{R}} + \lambda_{\text{M}} \cdot \mathcal{L}_{\text{M}} + \lambda_{\text{P}} \cdot \mathcal{L}_{\text{P}} + \lambda_{\text{reg}} \cdot \mathcal{L}_{\text{reg}},
\end{equation}
where $\lambda_{\text{M}}$, $\lambda_{\text{P}}$ and $\lambda_{\text{reg}}$ are scaling hyperparameters.
\section{Evaluation on Autoencoders}
\label{sec:exp_vmae}

In this section, we first evaluate our VMAEs in the aspect of latent space, comparing with other autoencoders introduced in \cref{sec:prelim:autoencoder}. 
All autoencoders are trained on the ImageNet-1K~\cite{deng2009imagenet} training set with their respective optimal hyperparameters using 8 NVIDIA A100 GPUs (40GB). Elaboration on datasets and implementation details can be found in \cref{apdx:autoencoder_setup}.

\begin{figure}[t]
    \centering
    \vspace{-0.1cm}
    \includegraphics[width=\linewidth]{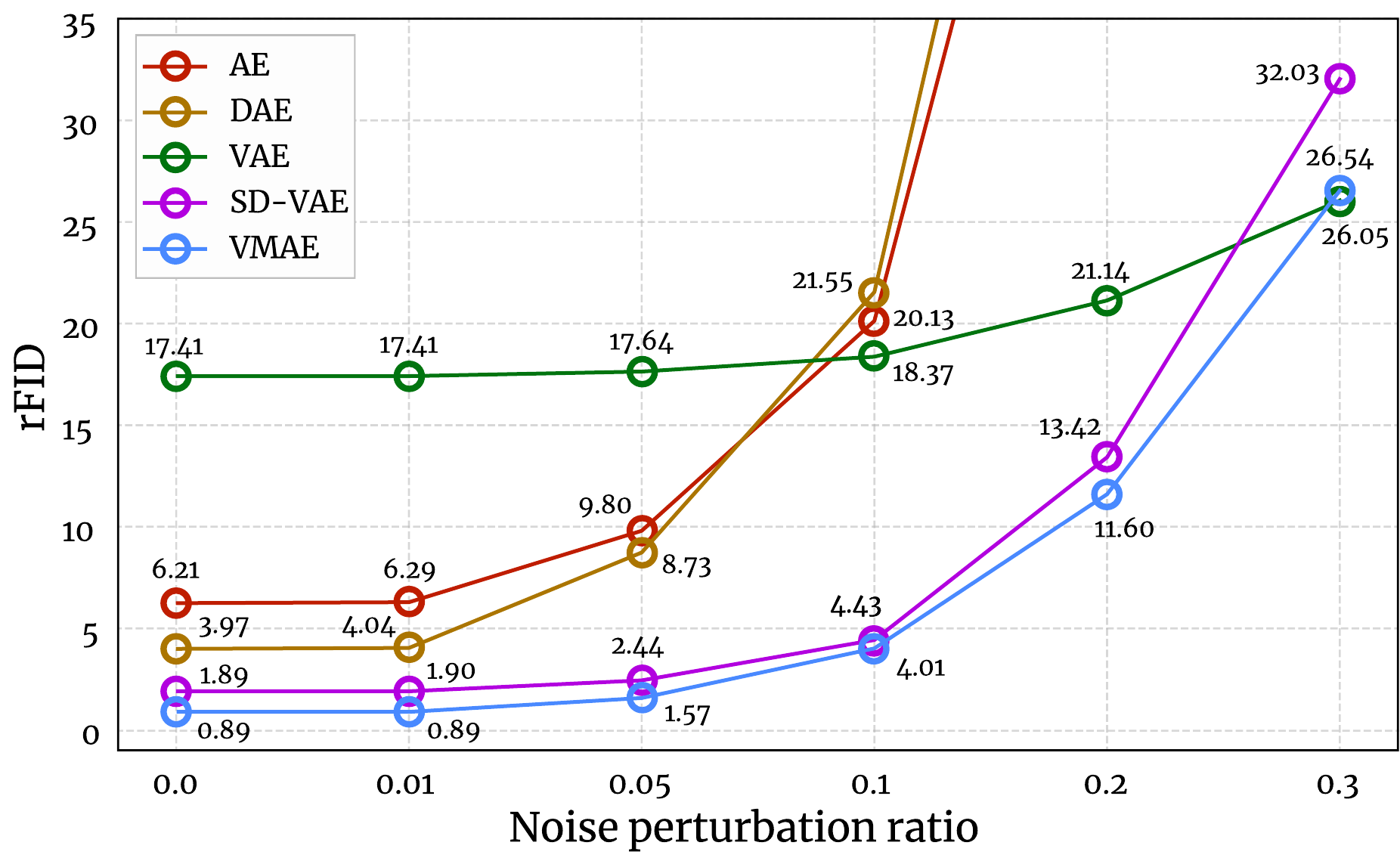}
    \vspace{-.5cm}
    \caption{
        \textbf{Robustness to the noise perturbation in latent space}. Probabilistic autoencoders (VAE, SD-VAE, VMAE) demonstrate greater robustness to latent perturbations, as their decoders can reconstruct from a smoother latent space compared to deterministic autoencoders (AE, DAE).
    }
    \label{fig:exp:smoothness}
    \vspace{-0.5cm}
\end{figure}

\subsection{Results and Analysis}
\label{sec:exp:ae_analysis}
\noindent
\textbf{Latent Space Smoothness.}
As introduced in \cref{method:smoothness}, the latent space should be smooth and continuous rather than consisting of sparse points to ensure stable and consistent score matching.
As illustrated in \cref{fig:support}, deterministic encoders (AE, DAE) project inputs into discrete and sparse latent \emph{points}, restricting the decodable space to the discrete isolated points.
In contrast, probabilistic encoders (VAE, SD-VAE, VMAE) map inputs to latent distributions, forming a continuous latent space that broadens the decodable region.
A continuous \emph{space} is preferable to discrete \emph{points} for diffusion models to learn, as it allows generations within the vicinity of target points rather than requiring exact matches.

To assess the smoothness of the latent space,
we compare the robustness of each autoencoder to a noise with varied perturbation ratios.
\cref{fig:exp:smoothness} reports rFID scores of each autoencoder with varied noise levels, measuring the quality of generated images.
First, the deterministic autoencoders (AE and DAE) show reasonably good performance when the noise level is extremely low, while the rFID score significantly drops when a larger level of perturbation is applied.
This suggests that deterministic encoding approaches would destabilize the diffusion model's learning process due to their limited support in the latent space.
In spite of their completely deterministic design, they still show robustness for a very small noise, probably because the training process causes some level of errors when they map images to the latent space and reconstruct them, and the model learns to robustly map within this small range of noise.
Second, VAE shows minimal increase in rFID across perturbation ratios, indicating that its decodable space is most widely supported.
Nevertheless, its high absolute rFID scores make it less suitable for LDMs.
Lastly, SD-VAE and VMAE demonstrate combined strengths, achieving the best quality (in rFID) while more robust to noise compared to AE and DAE.
This indicates
that a probabilistic latent space indeed results in a larger decodable space which is more desirable for LDMs.
Comparing the two, our VMAE consistently outperforms the strongest baseline, SD-VAE across all noise levels.
This conclusion is consistent with the actual generation quality, demonstrated in \cref{tab:diffusion_performance}.



\begin{figure}[t]
    \centering
    \includegraphics[width=\linewidth]{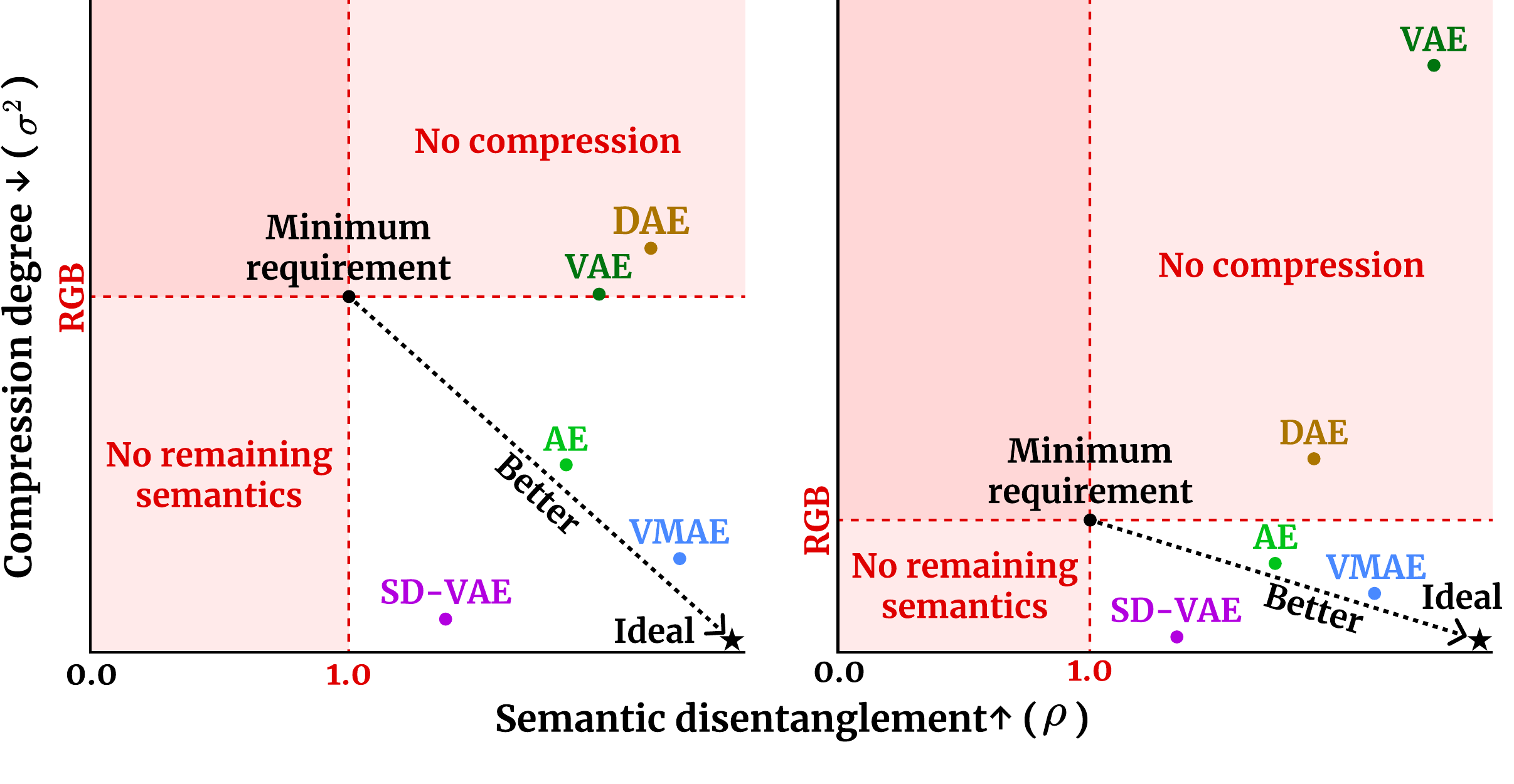}
    \raisebox{.35cm}{\footnotesize \text{~~~(a) Object-level ~~~~~~~~~~~~~~~~~~~~~~~~~~~~~~~~~~~~ (b) Part-level}} \\ 
    
    \vspace{-.3cm}
    \caption{\textbf{Perceptual compression evaluation of autoencoders.} (a) object-level and (b) part-level are evaluated with ADE20K and CelebAMask-HQ, respectively. A lower compression degree ($\sigma^2$) signifies stronger compression, whereas higher semantic disentanglement ($\rho$) reflects the amount of semantic information retained after compression. Our VMAE is positioned closest to the ideal autoencoder, indicating the most effective perceptual compression.}
    \label{fig:exp:compression_evaluation}
    \vspace{-0.5cm}
\end{figure}


\noindent
\textbf{Perceptual Compression.}
As detailed in \cref{method:compression}, perceptual compression consists of
two key properties: how much perceptual detail is compressed during encoding and how well semantic information is preserved after encoding.
For the former, the more perceptual details are discarded during the compression, the smaller the variance among the features within a same cluster will be.
To quantify this, we measure the average \emph{feature variance} within each semantic cluster, denoted by $\sigma^2$, based on the ground truth segmentation map.
In the context of LDMs, an effective autoencoder should exhibit a higher degree of compression than the raw image; that is, a smaller $\sigma^2$ is desired.

For the latter, on the other hand, if the compression effectively retains the semantics, the semantic clusters should be highly distinguishable (disentangled) from each other in the latent space.
Specifically, we first construct an affinity graph of latent features from each image and compute the mean intra-cluster edges ($u_\text{intra}$) and mean inter-cluster edges ($u_\text{inter}$) based on the ground truth. 
Then, we quantify the semantic disentanglement by their ratio $\rho = u_\text{intra} / u_\text{inter}$.
In the context of LDMs, it is desired to maintain semantically distinguishable features; that is, $\rho > 1$.

To consider various compression levels, as proposed in \cref{method:compression}, we use ADE20K for object-level evaluation and CelebAMask-HQ for part-level evaluation, where the \emph{semantics} refer to the objects (\emph{e.g.}, chair, table) in ADE20K, while the parts in CelebAMask-HQ (\emph{e.g.}, nose, ears), respectively.


As shown in \cref{fig:exp:compression_evaluation}, DAE and VAE exhibit larger $\sigma^2$ than the raw image, indicating that they are unable to perform compression—a fundamental requirement for perceptual compression, while AE, SD-VAE, and our VMAE successfully achieve it.
Among these, SD-VAE shows the strongest compression, indicated by the lowest $\sigma^2$.
However, its $\rho$ is lowest among all methods, implying that
its compression not just reduces perceptual details but also erases essential semantic information, resulting in excessive feature entanglement.
Notably, VMAE most successfully balances and simultaneously satisfies the two criteria, \emph{i.e.}, strong compression capability and superior semantic disentanglement.

\begin{figure}[t]
    \centering

    \includegraphics[width=\linewidth]{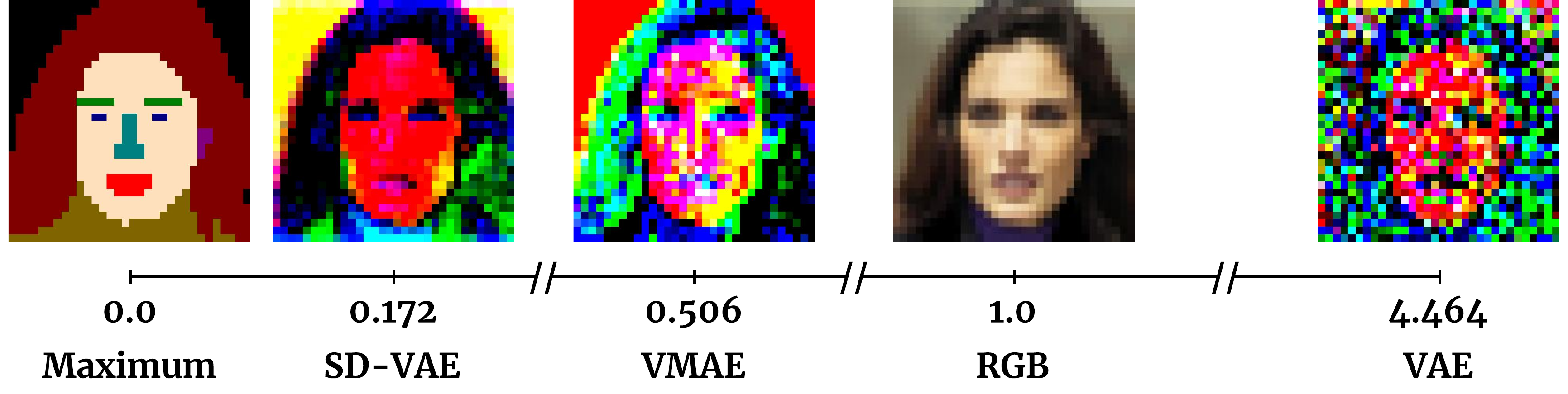}
    \raisebox{.15cm}{\footnotesize \text{(a) Part-level (CelebAMask-HQ)}} \\
    
    \vspace{.2cm}
    
    \includegraphics[width=\linewidth]{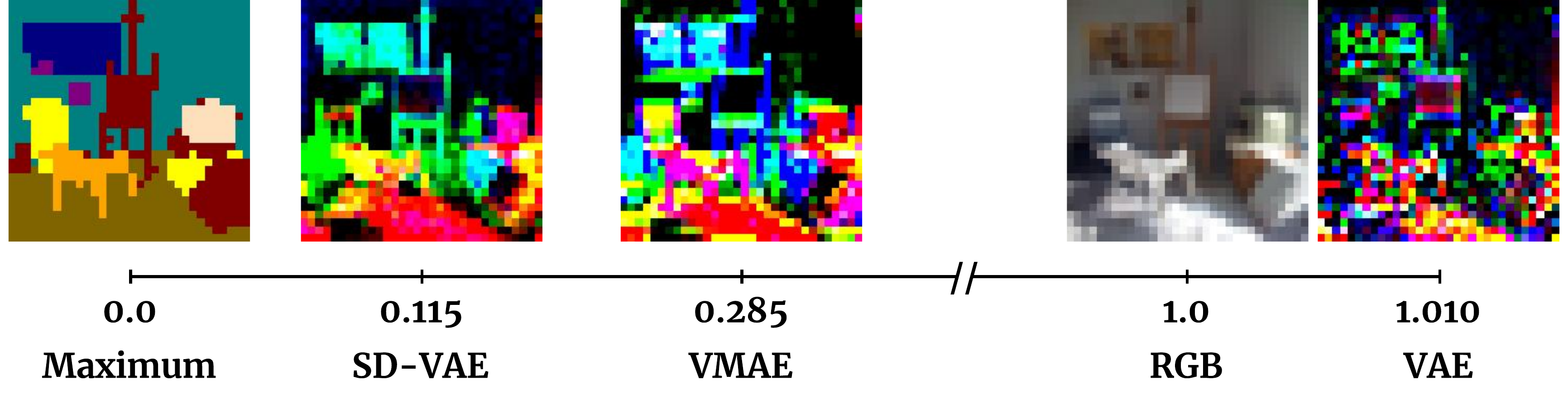}
    \raisebox{.15cm}{\footnotesize \text{(b) Object-level (ADE20K)}} \\ 

    \vspace{-.2cm}
    \caption{\textbf{Illustration of compressed latents of autoencoders, along with relative compression degrees ($\sigma^2$).} VMAE achieves a strong compression with diversified features, preserving more detailed semantics, unlike VAE with highly dispersed features or SD-VAE with excessively entangled features.
    }
    \vspace{-.6cm}
    \label{fig:exp:compression_visualization}
\end{figure}
For further qualitative analysis, we illustrate the compressed latents with respect to the degree of compression ($\sigma_\text{intra}$) in \cref{fig:exp:compression_visualization}.
As clearly shown in (a) part-level compression, VAE features are highly dispersed, implying that they are embedded closer to a Gaussian distribution rather than a semantically meaningful one.
This observation aligns with findings in $\beta$-VAE~\cite{higgins2017betavae}.
Also, as the objects in the same class are not distinctly labeled in ADE20K, the extremely low $\sigma^2$ of SD-VAE in (b) object-level compression suggests that its latent features are not only entangled within individual objects but also struggle to discriminate different objects.
This result strongly suggests that the latent features of SD-VAE are highly entangled, leading to degraded decoding performance (\cref{tab:reconstruction_quality}) and making them unsuitable for downstream tasks (\cref{method:compression}).
The superior performance of our VMAE is evident in both (a) and (b), where its features are diversified yet semantically structured, indicating its \emph{hierarchical} compression.

\begin{table}[t]
\centering
\caption{Reconstruction performance comparison across autoencoders on ImageNet-1K test set.}
\vspace{-.2cm}
\resizebox{\columnwidth}{!}{%
\begin{tabular}{l|cccccr}
\toprule
\textbf{Model} & \textbf{L1}$_\downarrow$ & \textbf{MSE}$_\downarrow$ & \textbf{PSNR}$_\uparrow$ & \textbf{SSIM}$_\uparrow$ & \textbf{LPIPS}$_\downarrow$ & \textbf{rFID}$_\downarrow$ \\
\midrule
AE & \textbf{0.0218} & \textbf{0.0013} & \textbf{32.18} & \textbf{0.895} & 0.172 & 6.21\\
DAE & 0.0237 & 0.0015 & 31.31 & 0.887 & 0.175 & 3.97\\
VAE & 0.0281 & 0.0018 & 29.40 & 0.825 & 0.281 & 17.41\\
SD-VAE & 0.0223 & 0.0015 & 29.85 & 0.853 & \underline{0.099} & \underline{1.89} \\
\textbf{VMAE (Ours)} & \underline{0.0221} & \underline{0.0014} & \underline{31.52} & \underline{0.890} & \textbf{0.062}& \textbf{0.89}\\
\midrule
Ground truth & 0.0000 & 0.0000 & \(\infty\) & 1.000 & 0.000 & 0.000\\
\bottomrule
\end{tabular}%
}
\vspace{-.2cm}
\label{tab:reconstruction_quality}
\end{table}

\begin{table}[t]
\centering
\caption{Model efficiency comparisons. The numbers in parentheses indicate the relative percentage of our method's efficiency.}
\vspace{-.2cm}
\resizebox{.9\columnwidth}{!}{%
\begin{tabular}{l|c|c}
\toprule
\textbf{Metric} & \textbf{AE, DAE, VAE, SD-VAE} & \textbf{VMAE (Ours)} \\
\midrule
\textbf{Model size (MB)} & 319.7 & \textbf{42.7 ($13.4\%$)} \\
\textbf{GFLOPS} & 17,331.3 & \textbf{703.9 ($4.1\%$)} \\
\textbf{Training time (hr)} & 24 & \textbf{9 ($37.5\%$)} \\
\bottomrule
\end{tabular}%
}
\vspace{-.4cm}
\label{tab:model_efficiency}
\end{table}

\begin{figure*}[t!]
    \centering
    \includegraphics[width=\textwidth]{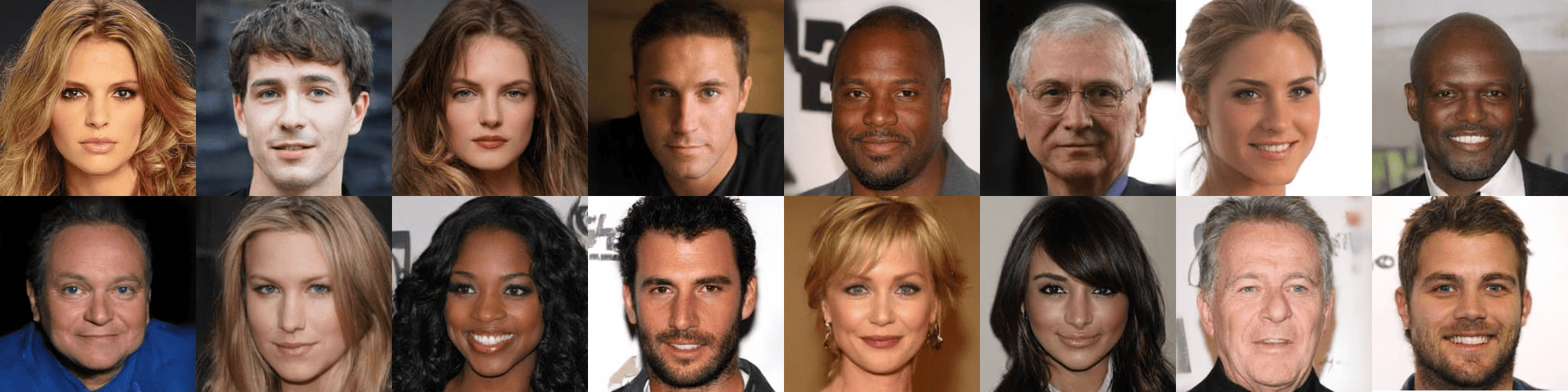}
    \caption{\textbf{Generated Samples on CelebA-HQ ($256 \times 256$).} Our LDMAE successfully captures diversity in age, ethnicity, hairstyle, and other characteristics, while also preserving fine details of human faces.}
    \label{fig:dm_qualitative}
\vspace{-0.4cm}
\end{figure*}


\noindent
\textbf{Reconstruction Quality.}
We evaluate the reconstruction capability of autoencoders using various metrics on the ImageNet-1K test set.
\cref{tab:reconstruction_quality} indicates that AE and DAE excel in pixel-level reconstruction (L1, MSE, PSNR, SSIM), while they exhibit weaker perceptual reconstruction (LPIPS, rFID).
VAE performs poorly across all reconstruction metrics, making it unsuitable for LDMs.
This is possibly due to the strong influence of KL divergence loss, which forces the decoder to reconstruct images from sampled latents near a Gaussian distribution, making reconstruction more challenging.
In contrast, SD-VAE achieves strong perceptual reconstruction while maintaining comparable pixel fidelity, owing to the additional reconstruction loss, \emph{i.e.}, adversarial loss, which alleviates the strong influence of latent regularization.
Our VMAE surpasses SD-VAE in both perceptual detail recovery and pixel-level accuracy, benefiting from its hierarchically structured latent features, \emph{i.e.}, diversified representations that facilitate more precise decoding. In our VMAE, patch drop during the encoding likely weakens the effect of latent regularization.

\noindent
\textbf{Model Efficiency.}
We compare the model cost in \cref{tab:model_efficiency}.
Our VMAE demonstrates significantly faster training and inference, with considerably smaller model size.
Faster inference is particularly favorable for training LDMs, as image encoding is required at each iteration.






\section{Evaluation on Overall Image Generation}
\label{sec:exp_ldm}

We conduct experiments to evaluate the performance of our LDMAE, an LDM model equipped with the proposed VMAE, for image generation. The datasets and detailed experimental setup is provided in \cref{apdx:generation_setup}.

\subsection{Results and Analysis} 
\label{sec:exp_ldm:result}


\cref{tab:diffusion_performance} presents the generation performance of LDMs equipped with different autoencoders. We can analyze these results through the lens of the three desirable properties of autoencoders for LDMs.

AE and DAE, which lack smooth latent spaces, show weaker generation performance than probabilistic models like SD-VAE and VMAE, which benefit from smoother latent representations. Among deterministic models, DAE performs slightly better than AE, indicating that DAE’s latent space is actually more robust~\cite{vincent2008dae}.
VAE achieves the smoothest latent space, but its poor reconstruction quality results in the worst generation scores on both datasets. This highlights that latent smoothness alone is insufficient without reliable reconstruction. Notably, the gap between VAE and deterministic models (AE, DAE) is much smaller on CelebA-HQ than on ImageNet-1K. 
This is likely because the ImageNet-1K dataset is often poorly preprocessed, leaving objects off-center or smaller, with large, noisy backgrounds. Such background noise reduces sensitivity to fine reconstruction errors, especially in perceptual metrics like FID. 
In contrast, CelebA-HQ images are tightly cropped, with large, centered faces and minimal background, making them more sensitive to small errors in the predicted latent codes, which can be resolved by a sufficiently smooth latent space.
SD-VAE balances all three properties well, achieving solid performance on both datasets. VMAE further improves each property and consistently achieves the best scores across all metrics. Moreover, VMAE’s superior pixel-level reconstruction quality, while not directly reflected in FID and IS, likely enhances the perceptual quality of generated images. \cref{fig:dm_teaser,fig:dm_qualitative} show generated samples from ImageNet-1K and CelebA-HQ, demonstrating high-resolution results with balanced quality across diverse objects and faces.

\begin{table}[t]
\centering
\caption{Comparison of generation performance among different autoencoders on ImageNet-1K ($256\times256$) and CelebA-HQ ($256\times256$). 
The best values are highlighted in \textbf{bold}, and the second-best values are \underline{underlined}.}
\vspace{-.2cm}
\setlength{\tabcolsep}{3pt}
{\footnotesize
\resizebox{\linewidth}{!}{%
    \begin{tabular}{l|rrrrr|rr}
    \toprule
    \multirow{2}{*}{\textbf{Model}}  & \multicolumn{5}{c|}{\textbf{ImageNet-1K}} & \multicolumn{2}{c}{\textbf{CelebA-HQ}} \\
    \cmidrule(lr){2-6} \cmidrule(lr){7-8}
    & \textbf{gFID}$_\downarrow$ & \textbf{sFID}$_\downarrow$ & \textbf{IS}$_\uparrow$ 
    & \textbf{Prec}$_\uparrow$ & \textbf{Rec}$_\uparrow$ 
    & \textbf{gFID}$_\downarrow$ & \textbf{sFID}$_\downarrow$ \\
    \midrule
    AE & 12.92  & 12.65 & 124.0 & 0.724 & 0.339 & 24.80 & 29.90 \\
    DAE & 8.60 & 12.12 & 160.3 & 0.797 & 0.402  & 21.42 & 22.90 \\
    VAE & 34.60 & 22.32 & 54.6 & 0.517 & 0.415 & 32.33 & 32.96 \\
    SD-VAE & \underline{6.49} & \underline{5.60} & \underline{173.3} & \underline{0.819} & \underline{0.429} & \underline{9.00} & \underline{14.83} \\
    \textbf{VMAE} & \textbf{5.98} & \textbf{5.16} & \textbf{185.5} & \textbf{0.844} & \textbf{0.435} & \textbf{7.61} & \textbf{9.08} \\
    \bottomrule
    \end{tabular}
}
}
\label{tab:diffusion_performance}
\vspace{-.5cm}
\end{table}

\subsection{Ablation Study}
\label{sec:exp:ablation}
To evaluate the impact of each loss term on reconstruction and generation quality, we conduct an ablation study on ImageNet-1K ($256\times256$).
\cref{tab:ablation_losses} presents the results across five metrics: rFID, PSNR, LPIPS, SSIM, and gFID.

Starting only with the MSE reconstruction loss, same as the standard AE, the baseline focuses solely on pixel-level fidelity.
Consequently, it excels in PSNR and SSIM but underperforms in perceptual metrics (rFID, LPIPS).
As analyzed in \cref{sec:exp:ae_analysis}, the AE lacks latent smoothness, ultimately leading to high gFID.
When the masking loss $\mathcal{L}_\text{M}$ is added, the model becomes similar to vanilla MAE.
Although this results in some loss of pixel-level reconstruction quality, it yields a more aligned latent space, thereby improving perceptual reconstruction.
As a result, the gFID also improves substantially, although latent smoothness remains inadequate.
Adding the latent regularizer $\mathcal{L}_\text{reg}$
loss 
leads to a slight drop in PSNR and SSIM but confers a substantial gain in perceptual reconstruction quality, which carries over to better generative performance.
Finally, adding the perceptual loss $\mathcal{L}_\text{P}$ further enhances perceptual reconstruction quality, not only slightly recovering pixel-level reconstruction but also significantly improving rFID and LPIPS, with a modest improvement in gFID.

\begin{table}[t]
\centering
\caption{\textbf{Ablation study on the effect of each loss term}. We report reconstruction quality using rFID, PSNR, LPIPS, and SSIM and assess generation quality using gFID on ImageNet-1K ($256\times256$).}
\vspace{-.2cm}
\setlength{\tabcolsep}{5pt}
\resizebox{\columnwidth}{!}{%
    \begin{tabular}{l|cccc|r}
        \toprule
        \textbf{Losses} & \textbf{PSNR} & \textbf{SSIM} & \textbf{rFID} & \textbf{LPIPS} & \textbf{gFID} \\
        \midrule
        Baseline (MSE)  & 32.18 & 0.906 & 6.21 & 0.172 & 12.92  \\
        $+$~Masking loss ($\mathcal{L}_\text{M}$) & 32.01 & 0.913  & 4.66 & 0.130 & 8.92 \\
        $+$~Latent regularizer ($\mathcal{L}_\text{reg}$) & 31.14 & 0.881  & 1.59 & 0.112 & 6.32 \\
        $+$~Perceptual loss ($\mathcal{L}_\text{P}$) & 31.52 & 0.889  & 0.89 & 0.062 & 5.98 \\
        \bottomrule
    \end{tabular}
    \label{tab:ablation_losses}
    }
\vspace{-.5cm}
\end{table}


\section{Summary}
\label{sec:summary}

We propose three key properties for autoencoders in LDMs: latent smoothness, perceptual compression, and high-quality reconstruction. By analyzing existing autoencoders, we introduce Variational Masked AutoEncoders (VMAEs) to address their limitations while balancing these properties. Extensive experiments show that VMAEs enhance generative performance with a stable training process and highly efficient inference. Our findings provide valuable insights for designing effective autoencoders for LDMs.



\section*{Acknowledgments}
This work was supported by Samsung Electronics (IO240512-09881-01), Youlchon Foundation, NRF grants (RS-2021-NR05515, RS-2024-00336576, RS-2023-0022663) and IITP grants (RS-2022-II220264, RS-2024-00353131) by the government of Korea.

{
    \small
    \bibliographystyle{ieeenat_fullname}
    \bibliography{ref}
}
\clearpage
\setcounter{page}{1}
\maketitlesupplementary

\appendix

\pagenumbering{roman}
\renewcommand\thetable{\Roman{table}}
\renewcommand\thefigure{\Roman{figure}}
\setcounter{table}{0}
\setcounter{figure}{0}

\section{Autoencoders}
\label{apdx:prelim:autoencoders}

\textbf{AutoEncoders (AEs)}~\cite{10.5555/104279.104293} are optimized by solely relying on a reconstruction loss, 
compressing the input into a more compact latent space and reconstruct it back.
For its objective, Mean Squared Error (MSE) is commonly used:
\begin{equation}
  \mathcal{L}_{\text{AE}} = \mathbb{E}_{p_{\text{data}}(\boldsymbol{x})} \Big[ \|\boldsymbol{x} - g_\theta(f_\phi(\boldsymbol{x}))\|^2 \Big],
\end{equation}
where $p_{\text{data}}(\boldsymbol{x})$ is the data distribution.
The encoder $f_\phi(\boldsymbol{x})$ maps the input into the latent $\boldsymbol{z}$, and the decoder $g_\theta(\boldsymbol{z})$ recovers the latent $\boldsymbol{z}$ back to the input.
$\phi$ and $\theta$ are the parameters of the encoder and decoder networks, respectively.

\noindent
\textbf{Denoising AutoEncoders (DAEs)}~\cite{vincent2008dae} is trained to recover the original input from the noised one for the robust latent features with improved generalization. The objective is
\begin{equation}
  \mathcal{L}_{\text{DAE}} = \mathbb{E}_{p_{\text{data}}(\boldsymbol{x}), p_c(\tilde{\boldsymbol{x}}|\boldsymbol{x})} \Big[ \|\boldsymbol{x} - g_\theta(f_\phi(\tilde{\boldsymbol{x}}))\|^2 \Big],
\end{equation}
where $p_c(\tilde{\boldsymbol{x}}|\boldsymbol{x})$ represents a corrupted data distribution. 

\noindent
\textbf{Variational AutoEncoders (VAEs)}~\cite{kingma2013vae} adopt a probabilistic framework by encoding the input data into a latent variable distribution instead of a fixed vector, facilitating sampling.
VAEs are trained using the Evidence Lower Bound Loss (ELBO), which combines a reconstruction loss with a prior matching term, \emph{i.e.}, KL-divergence to regularize the latent space towards a Gaussian distribution:
\begin{equation}
\begin{aligned}
  \mathcal{L}_{\text{VAE}} = \mathbb{E}_{p_{\text{data}}(\boldsymbol{x})} \Big[ \underset{\text{reconstruction}} {\underbrace{\mathbb{E}_{q_{\phi}(\boldsymbol{z}|\boldsymbol{x})}[-\log p_\theta(\boldsymbol{x}|\boldsymbol{z})]}} \\
  &\hspace*{-3cm} + \lambda_{\text{KL}} \cdot \underset{\text{prior matching}}{\underbrace{\text{D}_\text{KL}(q_\phi(\boldsymbol{z}|\boldsymbol{x}) \| p(\boldsymbol{z}))}} \Big],
\end{aligned}
\end{equation}
where $q_\phi(\boldsymbol{z}|\boldsymbol{x})$ is the distribution of latent $\boldsymbol{z}$ encoded from the input $\boldsymbol{x} \sim p_{\text{data}}$ and $p_\theta(\boldsymbol{x}|\boldsymbol{z})$ is the distribution of reconstructed $\boldsymbol{x}$ from $\boldsymbol{z}$. KL-divergence is scaled by $\lambda_{\text{KL}}$.

\noindent
\textbf{StableDiffusion VAEs (SD-VAEs)}~\cite{rombach2022ldm,esser2024sd3} are built upon the VQGAN~\cite{esser2021vqgan}, the autoencoder adopted in traditional LDMs~\cite{rombach2022ldm}. 
Following the VQGAN, SD-VAEs integrate an additional adversarial network and train with 
perceptual loss (LPIPS)~\cite{zhang2018lpips} for an improved perceptual quality in the compressed space.
Unlike VQGAN, however, SD-VAEs omit the quantization layer entirely; instead, it simply adopts continuous features.
In this paper, we follow the SD-VAEs settings from StableDiffusion3~\cite{esser2024sd3}, unless noted otherwise.
The overall objective of SD-VAEs is
\begin{align}
  \mathcal{L}_{\text{SD-VAE}} &= \mathbb{E}_{p_{\text{data}}(\boldsymbol{x})} \Big[ \underset{\text{reconstruction}} {\underbrace{\mathbb{E}_{q_{\phi}(\boldsymbol{z}|\boldsymbol{x})}[-\log p_\theta(\boldsymbol{x}|\boldsymbol{z})]}} \\
  &\hspace*{2cm} + \lambda_{\text{KL}} \cdot \underset{\text{prior matching}}{\underbrace{\text{D}_\text{KL}(q_\phi(\boldsymbol{z}|\boldsymbol{x}) | p(\boldsymbol{z}))}} \Big] \nonumber \\
  &\hspace*{-1.5cm} + \lambda_{\text{D}} \cdot \underset{\text{adversarial loss}}{\underbrace{  \left( \mathbb{E}_{p_{\text{data}}}[\log D(\boldsymbol{x})] + \mathbb{E}_{q_\phi(\boldsymbol{z}|\boldsymbol{x})}[\log (1 - D(p_\theta(\boldsymbol{x}|\boldsymbol{z})))] \right)}} \nonumber \\
  &\hspace*{-1.5cm} + \lambda_{\text{LPIPS}} \cdot \underset{\text{LPIPS loss}} {\underbrace{\mathbb{E}_{p_{\text{data}}(\boldsymbol{x}), q_\phi(\boldsymbol{z}|\boldsymbol{x}), p_\theta(\hat{\boldsymbol{x}}|\boldsymbol{z})} \bigg[ \sum_{l} w_l \|\psi_l(\boldsymbol{x}) - \psi_l(\hat{\boldsymbol{x}})\|_2^2 \bigg]}}. \nonumber
\end{align}
The third term corresponds to the adversarial loss scaled by $\lambda_{\text{D}}$, where $D(x)$ denotes the discriminator function.
The last LPIPS loss term incorporates the feature extraction function $\psi_l$ up to $l$ layers of a pre-trained network, with $w_l$ as the layer-specific weight, scaled by $\lambda_{\text{LPIPS}}$.

\noindent
\textbf{Masked AutoEncoders (MAEs)}~\cite{he2022mae} were originally proposed as a self-supervised learning method for representation learning based on Vision Transformers (ViTs)~\cite{dosovitskiy2020vit}.
The MAEs encoder $f_\phi(\boldsymbol{x}_v)$ maps a masked-out image $\boldsymbol{x}_v \sim p(\boldsymbol{x}_v|\boldsymbol{x})$ into a latent $\boldsymbol{z}$, and its decoder $g_\theta(\boldsymbol{z})$ reconstructs the original input $\boldsymbol{x} \sim p_{\text{data}}(\boldsymbol{x})$ from $\boldsymbol{z}$ along with learnable mask tokens.
Since MSE loss applies only to mask tokens, the actual loss acts more as a prediction loss than a reconstruction loss.
Omitting the mask tokens for simplicity, the objective can be expressed as
\begin{equation}
  \mathcal{L}_{\text{MAE}} = \mathbb{E}_{p_{\text{data}}(\boldsymbol{x}), p(\boldsymbol{x}_v|\boldsymbol{x})} \Big[ \|\mathrm{M} \odot (\boldsymbol{x} - g_\theta(f_\phi(\boldsymbol{x}_v)))\|^2 \Big],
\end{equation}
where $\mathrm{M}$ is a fixed-ratio random binary mask.

\section{Implementation Details}
\subsection{Experimental Setup for Autoencoders}
\label{apdx:autoencoder_setup}
\textbf{Datasets.}
We use ImageNet-1K~\cite{deng2009imagenet} training set for training autoencoders, and evaluate them on the ImageNet-1K test set, ADE20k~\cite{zhou2019ade20k} test set, and CelebAMask-HQ~\cite{lee2020maskgan}. ADE20K and CelebAMask-HQ are segmentation datasets containing ground truth masks in pixel-level.

\noindent
\textbf{Implementation Details.}
Our VMAE adopts a symmetric ViT‐based encoder–decoder architecture.
The encoder partitions an input image $\mathbf{x} \in \mathbb{R}^{H \times W \times 3}$ into patch tokens ${\mathbf{x}_i \in \mathbb{R}^{h \times w \times d}}$ with patch size $(h,w)=(32,32)$ and embedding dimension $d=192$.
A masking ratio of 0.6 is applied, and the visible patches are processed by 12 Transformer layers.
The decoder prepends learnable mask tokens to the encoded sequence and reconstructs the full image using another 12 Transformer layers.
We impose a KL‐divergence loss on the latent representations and employ both pixel‐wise reconstruction loss and perceptual loss on the decoder outputs.

We train all autoencoders using the optimal hyperparameters for each model on 8 NVIDIA A100 GPUs (40GB).
The base learning rate is set to $10^{-5}$ for convolution-based models (AEs, DAEs, VAEs, and SD-VAEs), while $10^{-4}$ for our VMAEs.
Global batch size is set to 2048 except for the SD-VAEs, which require smaller batch size (256) for stable training of the adversarial network, following the implementation in VQGAN~\cite{esser2021vqgan}.

\begin{figure*}[t]
    \centering
    \includegraphics[width=\textwidth]{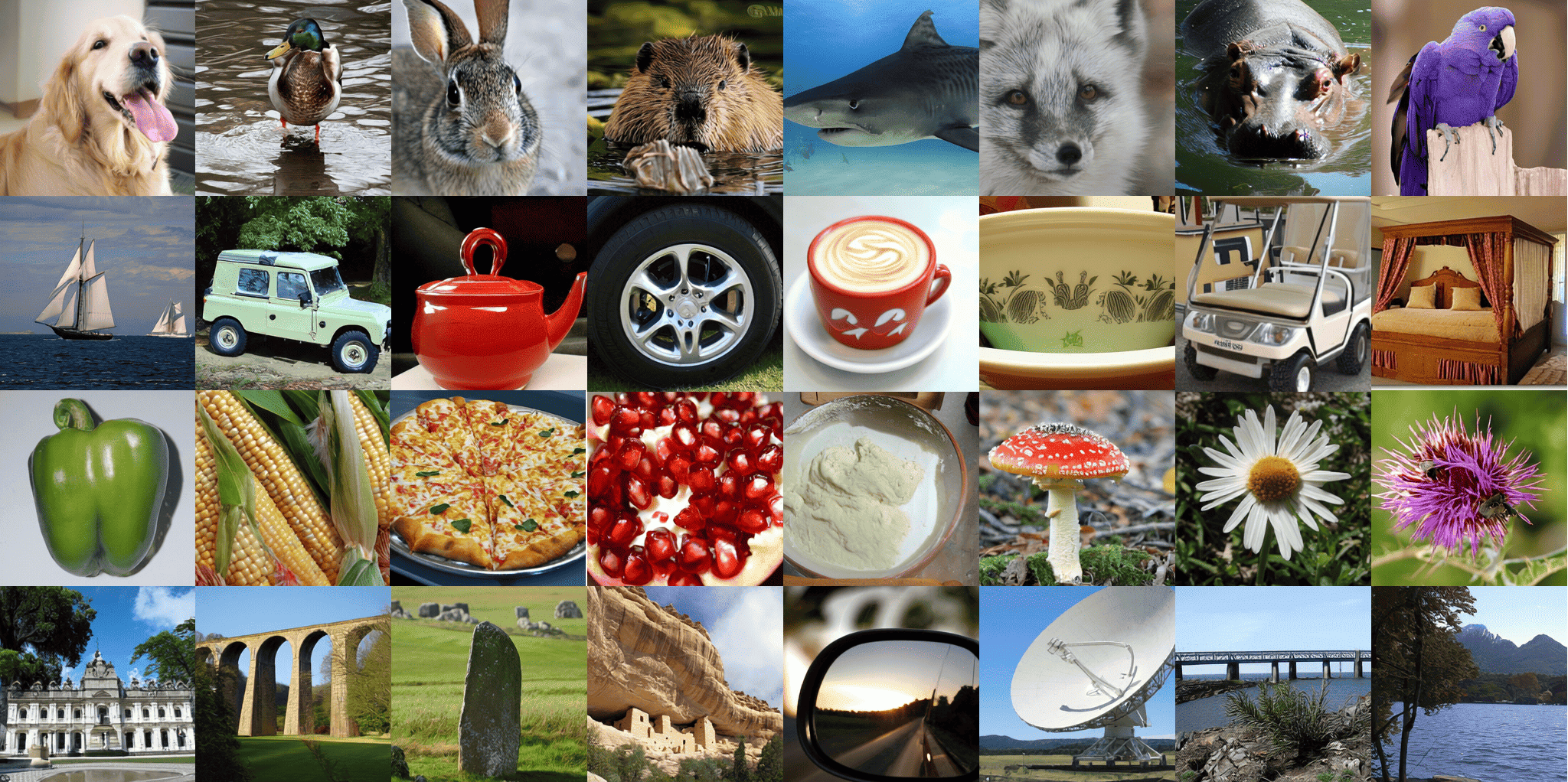}
    \caption{\textbf{Additional Class-Conditional Generation Examples on ImageNet-1K.} We present additional examples of class-conditional generation on ImageNet-1K ($256\times256$) across various classes.}
    \label{fig:dm_teasor}
\end{figure*}

\subsection{Experimental Setup for Image Generation}
\label{apdx:generation_setup}

\textbf{Datasets.}
We select the following datasets to test various aspects of generation performance.
For unconditional image generation, we use $256\times256$ downscaled 
CelebA-HQ~\cite{Karras2018celebahq}, a collection of 30,000 high-quality celebrity face images, commonly used for assessing face generation tasks.
For class-conditioned image generation, we train the diffusion model on ImageNet-1K~\cite{deng2009imagenet}, a large-scale dataset containing over 1.2 million labeled images on 1,000 categories, providing a rigorous benchmark.

\noindent
\textbf{Implementation Details.}
We employ DiT-B/1~\cite{peebles2023dit} as the diffusion model across all datasets, maintaining fixed hyperparameters for each dataset to ensure fair comparisons among autoencoders.
The learning rate is set to $2 \times 10^{-4}$ and the global batch size is fixed to 1024 for all datasets.
To mitigate divergence caused by uncontrolled attention logit growth, we apply QK normalization~\cite{dehghani2023scaling}.
We train for 100K iterations on ImageNet and 60K on CelebA-HQ.
For all other configurations, we use the default settings in~\citet{vavae}.
During sampling, we use a 250-step Euler integrator and apply classifier-free guidance (CFG) with a consistent scale across all architectures for class-conditional generation.

\noindent
\textbf{Evaluation Metrics.}
Inception Score (IS)~\cite{NIPS2016_is} measures how well a model captures the full class distribution while producing convincing class-specific samples.
Generative Fréchet Inception Distance (gFID)~\cite{heusel2017gans} calculates the distance between two image distributions in the Inception-v3 \cite{szegedy2016inception_v3} latent space, capturing both fidelity and diversity, and is widely regarded as more consistent with human judgment.
sFID~\cite{nash2021sfid}, a variation of FID using spatial features, better captures spatial relationships and high-level structure in image distributions.
Improved Precision and Recall~\cite{kynkaanniemi2019improved} both assess the fidelity of generated samples in different ways. Specifically, precision measures how realistic or high-quality the generated samples are, while recall evaluates whether the model captures the full diversity of the real dataset.

\begin{figure*}[t]
    \centering
    \begin{minipage}{0.48\linewidth}
        \centering
        \includegraphics[width=\linewidth]{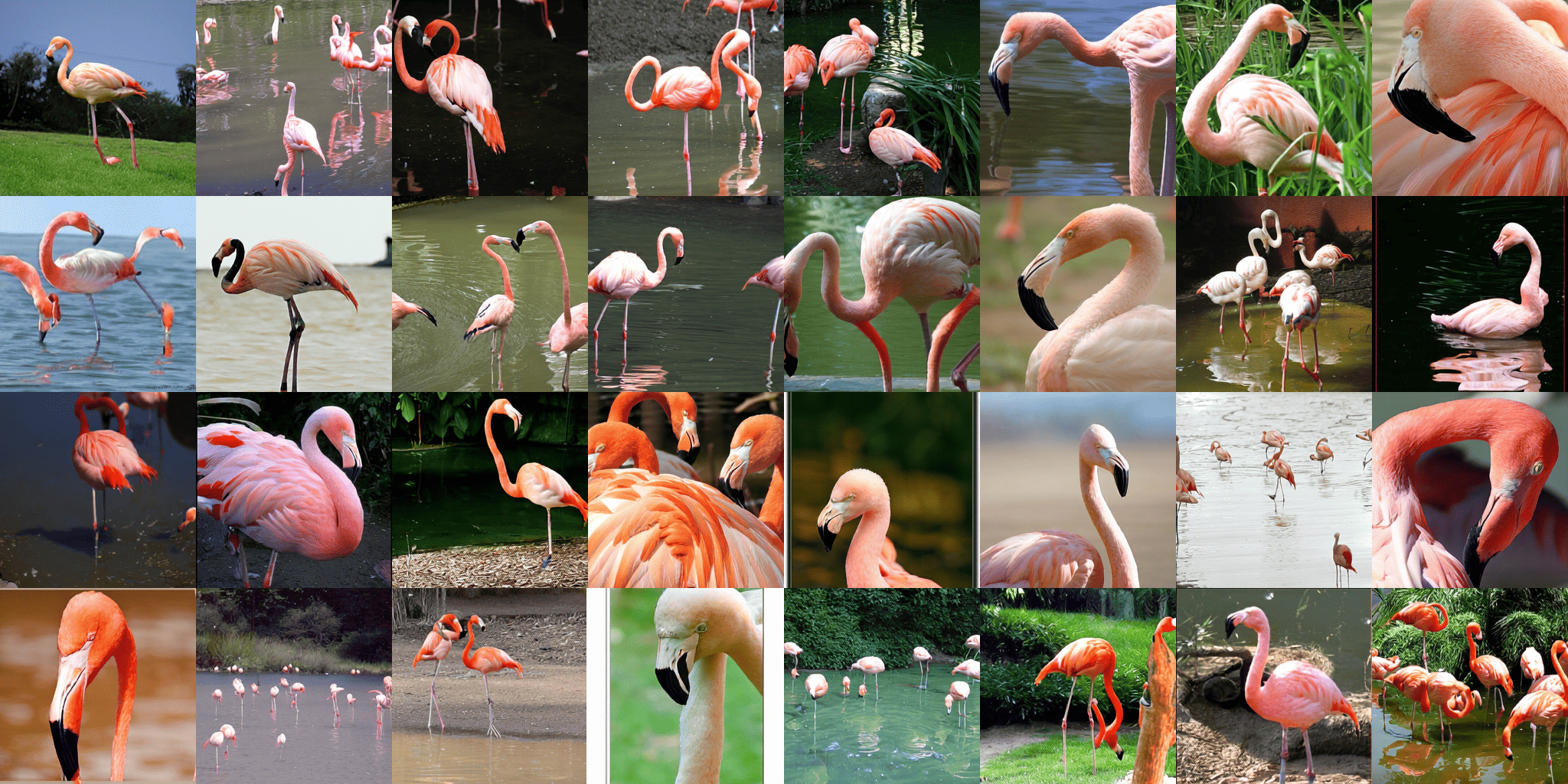}
        \caption*{(a) Flamingo (130)}
    \end{minipage}
    \hfill
    \begin{minipage}{0.48\linewidth}
        \centering
        \includegraphics[width=\linewidth]{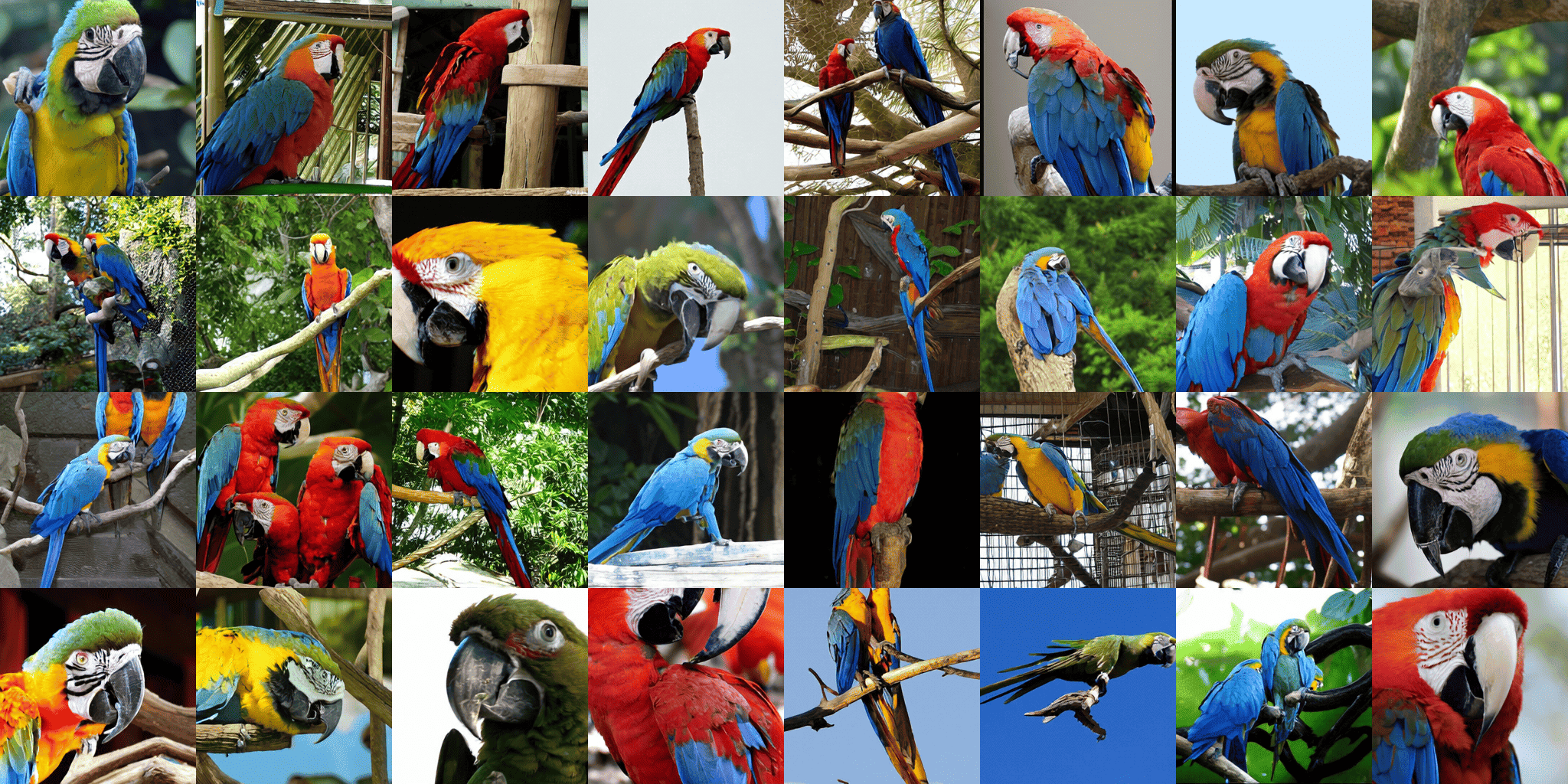}
        \caption*{(b) Macaw (88)}
    \end{minipage}
    
    \vspace{0.5em} 
    
    \begin{minipage}{0.48\linewidth}
        \centering
        \includegraphics[width=\linewidth]{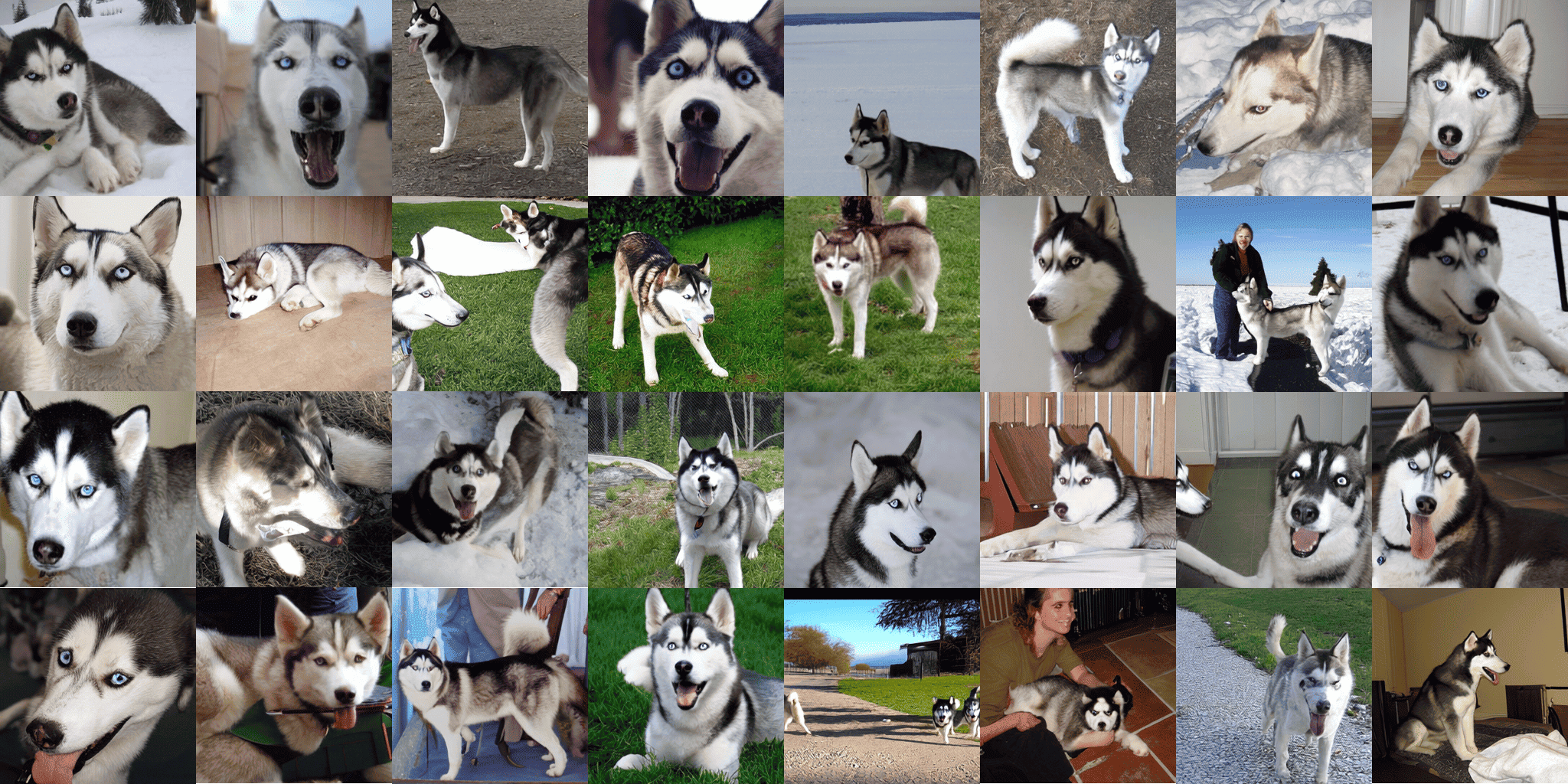}
        \caption*{(c) Siberian Husky (250)}
    \end{minipage}
    \hfill
    \begin{minipage}{0.48\linewidth}
        \centering
        \includegraphics[width=\linewidth]{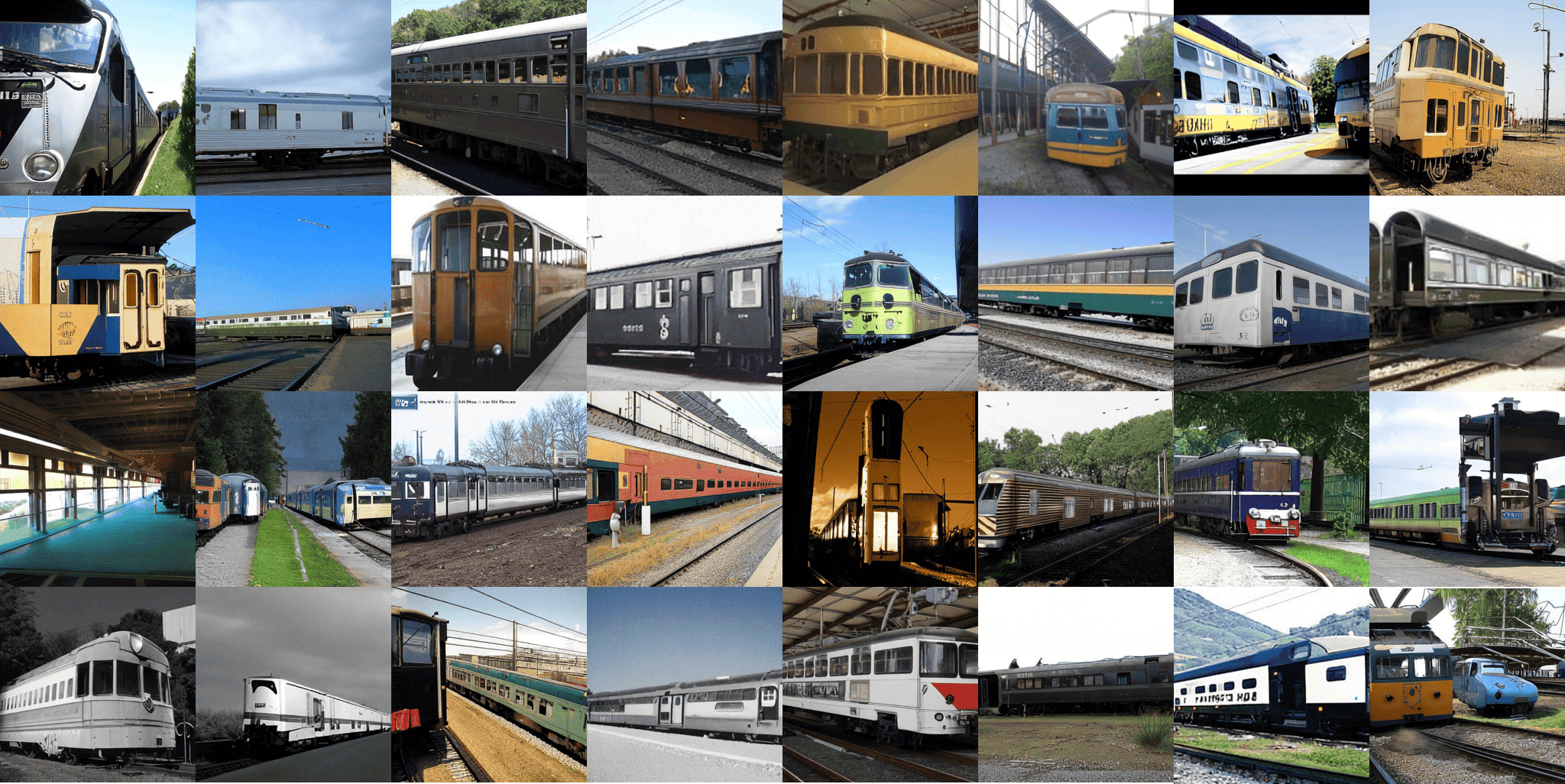}
        \caption*{(d) Passenger Car (705)}
    \end{minipage}
    
    \vspace{0.5em} 
    
    \begin{minipage}{0.48\linewidth}
        \centering
        \includegraphics[width=\linewidth]{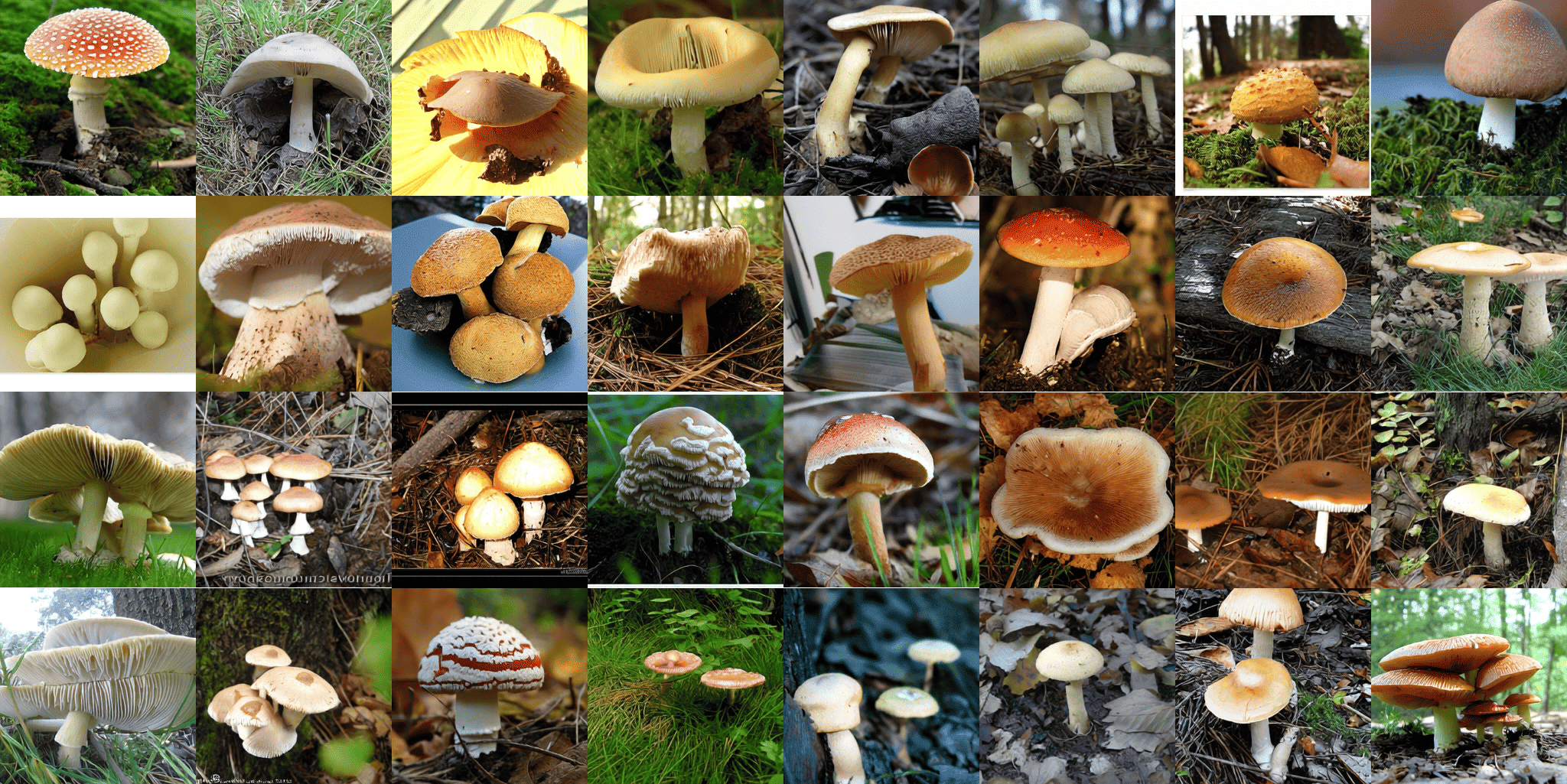}
        \caption*{(e) Mushroom (947)}
    \end{minipage}
    \hfill
    \begin{minipage}{0.48\linewidth}
        \centering
        \includegraphics[width=\linewidth]{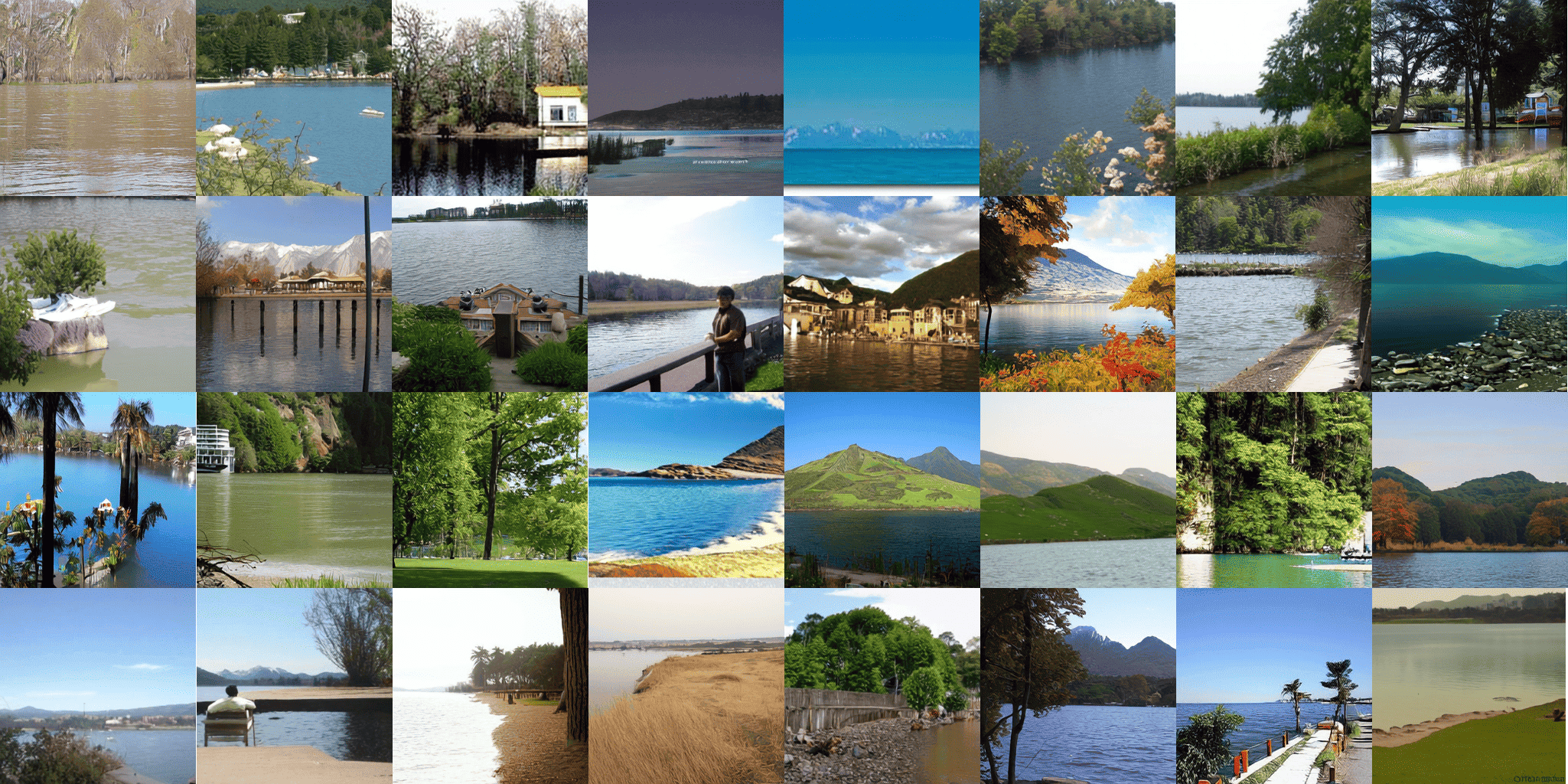}
        \caption*{(f) Lakeside (975)}
    \end{minipage}
    
    \vspace{0.5em} 
    
    \begin{minipage}{0.48\linewidth}
        \centering
        \includegraphics[width=\linewidth]{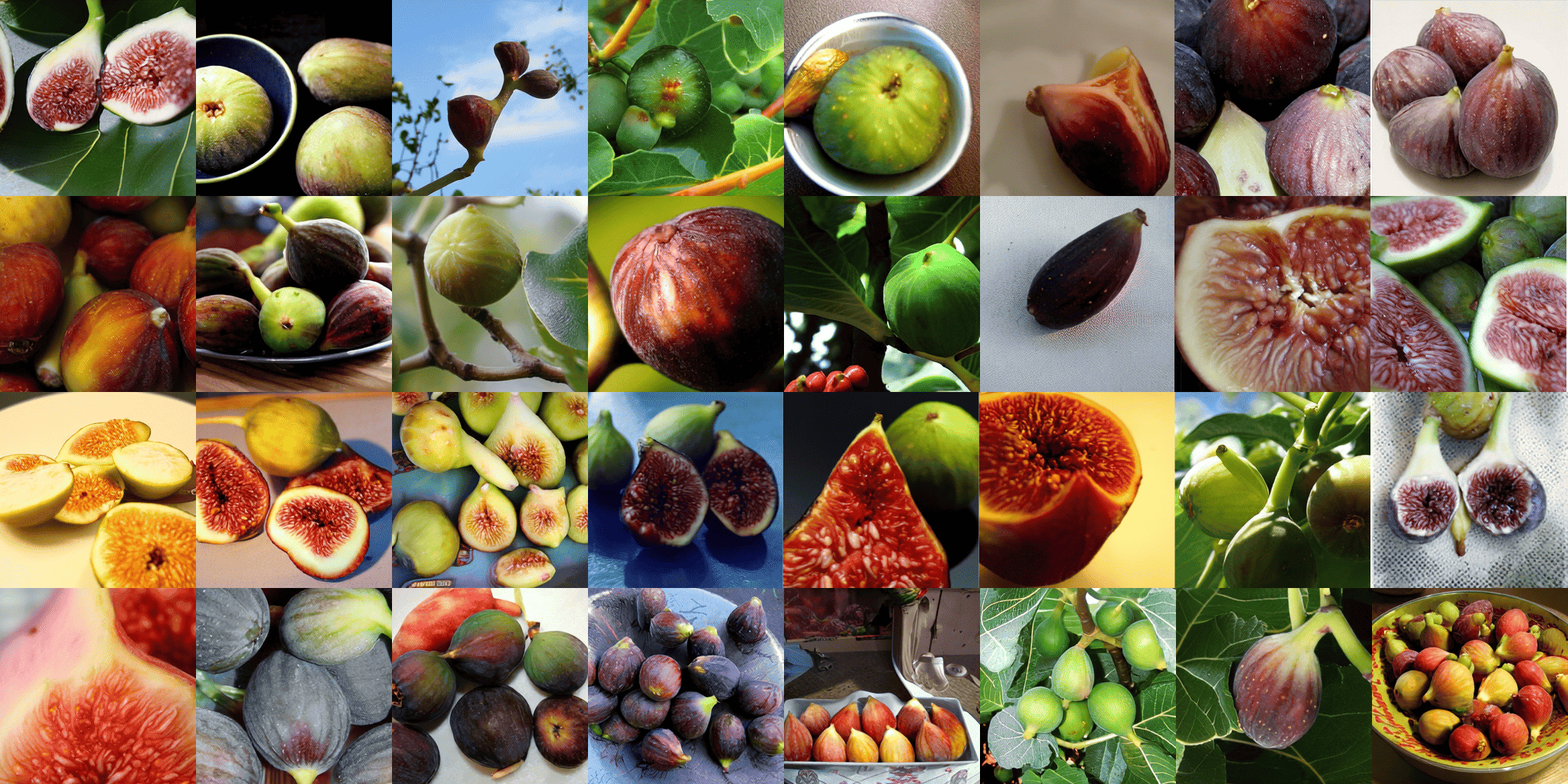}
        \caption*{(g) Fig (952)}
    \end{minipage}
    \hfill
    \begin{minipage}{0.48\linewidth}
        \centering
        \includegraphics[width=\linewidth]{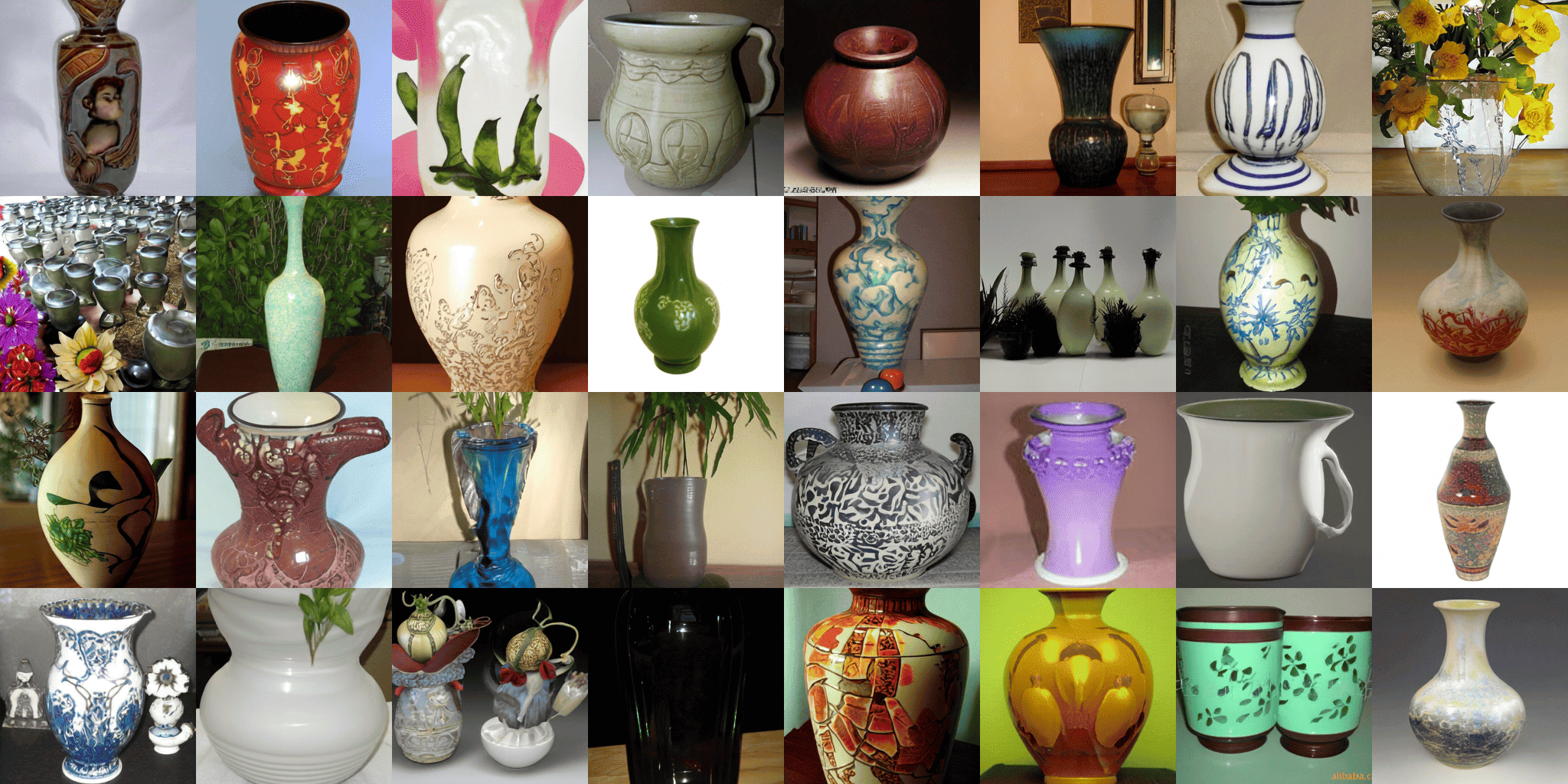}
        \caption*{(h) Vase (883)}
    \end{minipage}

    \caption{\textbf{Uncurated Class-Conditional Generation on ImageNet-1K.} We present a collection of uncurated class-conditional generation examples on ImageNet-1K at a resolution of $256\times256$. Each subcaption indicates the class name along with the corresponding class index.}
    \label{fig:uncurated1}
\end{figure*}



\end{document}